\documentclass[runningheads]{llncs}

\usepackage{eccv}

\usepackage{eccvabbrv}

\usepackage{graphicx}
\usepackage{booktabs}
\usepackage{rotating}

\usepackage[accsupp]{axessibility}  

\usepackage[pagebackref,breaklinks,colorlinks,citecolor=eccvblue]{hyperref}

\usepackage{orcidlink}

\begin{document}

\title{Fourier-Attentive Representation Learning: A Fourier-Guided Framework for Few-Shot Generalization in Vision-Language Models} 

\titlerunning{Fourier-Attentive Representation Learning}

\author{Hieu Dinh Trung Pham \and
Huy Minh Nhat Nguyen \and
Cuong Tuan Nguyen}

\authorrunning{Pham et al.}

\institute{Vietnamese German University\\
	Ho Chi Minh City, Vietnam\\
	{\tt\small
		\{104240027,\,10423045\}@student.vgu.edu.vn, 
		cuong.nt2@vgu.edu.vn
}}

\maketitle

\begin{abstract}
	Large-scale pre-trained Vision-Language Models (VLMs) have demonstrated strong few-shot learning capabilities. However, these methods typically learn holistic representations where an image's domain-invariant structure is implicitly entangled with its domain-specific style. This presents an opportunity to further enhance generalization by disentangling these visual cues. In this paper, we propose \textbf{F}ourier-\textbf{A}ttentive \textbf{R}epresentation \textbf{L}earning (FARL), a novel framework that addresses this by explicitly disentangling visual representations using Fourier analysis. The core of our method is a dual cross-attention mechanism, where learnable representation tokens separately query an image's structural features (from the phase spectrum) and stylistic features (from the amplitude spectrum). This process yields enriched, disentangled tokens that are then injected deep into the VLM encoders to guide adaptation. Our design, which includes an asymmetric injection strategy, forces the model to learn a more robust vision-language alignment. Extensive experiments on 15 datasets demonstrate the effectiveness of our approach.
\end{abstract}    
\section{Introduction}
\label{sec:intro}
Vision-Language Models (VLMs), such as CLIP~\cite{clip}, have shown remarkable zero-shot and few-shot transfer capabilities by aligning images and text in a shared embedding space. To adapt these foundation models efficiently, recent work has focused on prompt learning~\cite{coop, cocoop} and adapter-based strategies~\cite{maple, tcp, mma, mmrl}, which freeze the backbone and optimize lightweight learnable components. While effective, these methods remain fragile in low-data regimes.

We argue that this fragility stems not merely from limited capacity, but from a fundamental \textbf{spectral bias} in few-shot adaptation. It is well known that neural networks tend to rely on superficial statistics such as texture and color before learning robust semantic structure~\cite{geirhos2022imagenettrainedcnnsbiasedtexture}. From a Fourier perspective, these superficial cues are primarily encoded in the amplitude spectrum, while semantic structure and geometry are preserved in the phase spectrum~\cite{Oppenheim1980TheIO}. Existing adaptation methods operate on holistic feature embeddings where these spectral components are implicitly entangled, causing few-shot learners to overfit domain-specific amplitude statistics and limit their generalization capability under domain shift or to novel classes.

This exposes a gap in current approaches. Prompt learning methods treat visual representations as black boxes, offering no control over which spectral components drive adaptation. Meanwhile, Fourier-based techniques explored in domain generalization~\cite{xu_fourier-based_2021, yang2020fdafourierdomainadaptation} primarily use spectral manipulation as a data augmentation strategy, rather than as an explicit mechanism for representation learning within VLMs.

To bridge this gap, we propose \textbf{Fourier-Attentive Representation Learning (FARL)}, a framework that explicitly mitigates spectral bias during VLM adaptation. FARL decomposes images into phase-only (structure) and amplitude-only (style) components via the Fast Fourier Transform, and employs a dual cross-attention mechanism to disentangle and fuse these spectral cues. The resulting representations are injected into the VLM through a novel asymmetric strategy, enabling the model to align structure-centric visual features with semantically adaptive textual prompts while preserving robust visual representations.

Our contributions are summarized as follows:
\begin{itemize}
	\item We reframe few-shot VLM adaptation failure through the lens of spectral bias, showing that holistic adapters tend to overfit domain-specific amplitude statistics.
	\item We propose FARL, one of the first prompt learning frameworks to integrate Fourier-based disentanglement directly into the representation learning loop.
	\item Extensive experiments on 15 datasets demonstrate consistent improvements in base-to-novel generalization and cross-dataset transfer.
\end{itemize}

\section{Related Work}
\label{sec:related}

\subsection{Vision-Language Models and Efficient Adaptation:}
Large-scale Vision-Language Models (VLMs) like CLIP~\cite{clip}, ALIGN~\cite{align}, and FILIP~\cite{filip} have established a new paradigm in computer vision by aligning visual and textual modalities in a shared embedding space. While these models demonstrate impressive zero-shot capabilities, adapting them to specific downstream tasks with limited data remains a challenge. To address this, \textbf{Prompt Learning} methods emerged as a parameter-efficient solution, optimizing continuous input vectors rather than fine-tuning the entire backbone. Pioneering works like CoOp~\cite{coop} and CoCoOp~\cite{cocoop} focused on text-side prompting, while recent approaches like MaPLe~\cite{maple}, MMA~\cite{mma}, and MMRL~\cite{mmrl} introduced deep-layer adaptation across both modalities.

However, a limitation of these methods is their treatment of visual features as holistic embeddings. They lack explicit mechanisms to control \textit{what} information the model adapts to, particularly in terms of frequency-specific statistics. Consequently, in few-shot regimes, these adapters often overfit to domain-specific statistics (e.g., background textures or lighting conditions) implicitly entangled within the pre-trained features, rather than focusing on the robust structural semantics required for generalization.

\subsection{Fourier Analysis and Frequency Bias in Deep Learning:}
The motivation for our spectral disentanglement stems from a growing body of literature analyzing the \textbf{spectral bias} of neural networks. It has been observed that Deep Neural Networks (DNNs) often exhibit a tendency to learn "shortcuts" based on low-level, high-frequency statistics rather than high-level semantic shapes~\cite{geirhos2022imagenettrainedcnnsbiasedtexture, wang2020highfrequencycomponenthelps}. Geirhos et al.~\cite{geirhos2022imagenettrainedcnnsbiasedtexture} demonstrated that ImageNet-trained CNNs are heavily biased towards texture (amplitude-correlated features) over shape (phase-correlated features). In the context of few-shot adaptation, we argue that this bias is exacerbated: with scarce supervision, the adaptation modules eagerly lock onto the dominant amplitude statistics of the support set, leading to poor generalization on novel classes where those statistics shift.

Unlike prior work that merely analyzes this bias or attempts to mitigate it via training recipes, \textbf{FARL} integrates this insight directly into the architecture. We use Fourier decomposition not just as a preprocessing step, but as a mechanism to \textit{steer} the representation learning process, forcing the adaptation modules to explicitly query and attend to the information that is more correlated with domain-invariant structural semantics.

\subsection{Fourier-based Domain Generalization and Its Limitations for VLM Adaptation:}

\begin{figure}[t]
	\centering
	\begin{subfigure}{\linewidth}
		\centering
		\includegraphics[width=0.19\linewidth]{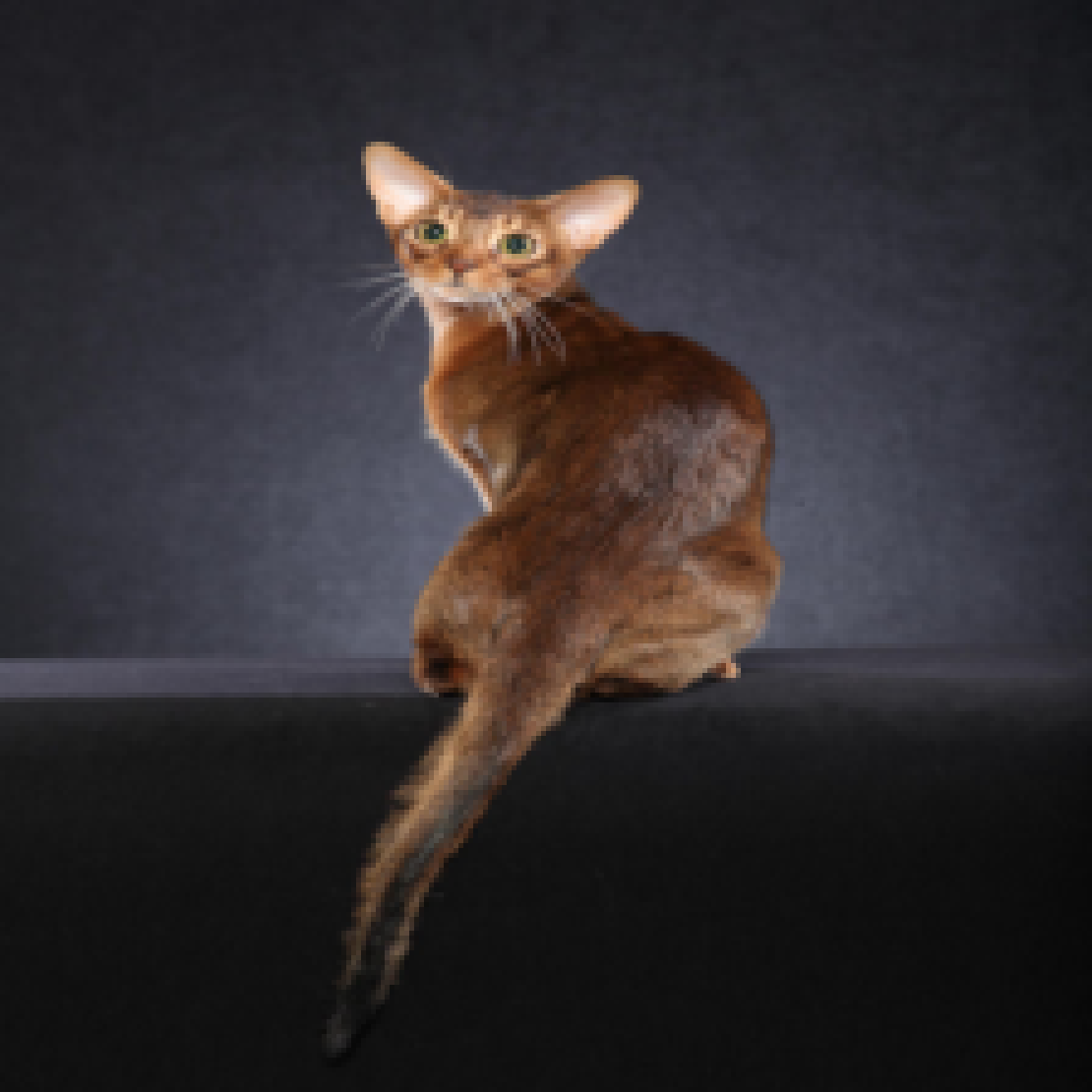}
		\includegraphics[width=0.19\linewidth]{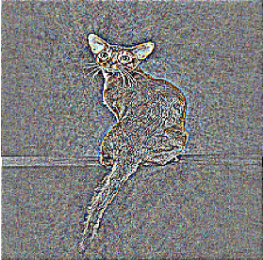}
		\includegraphics[width=0.19\linewidth]{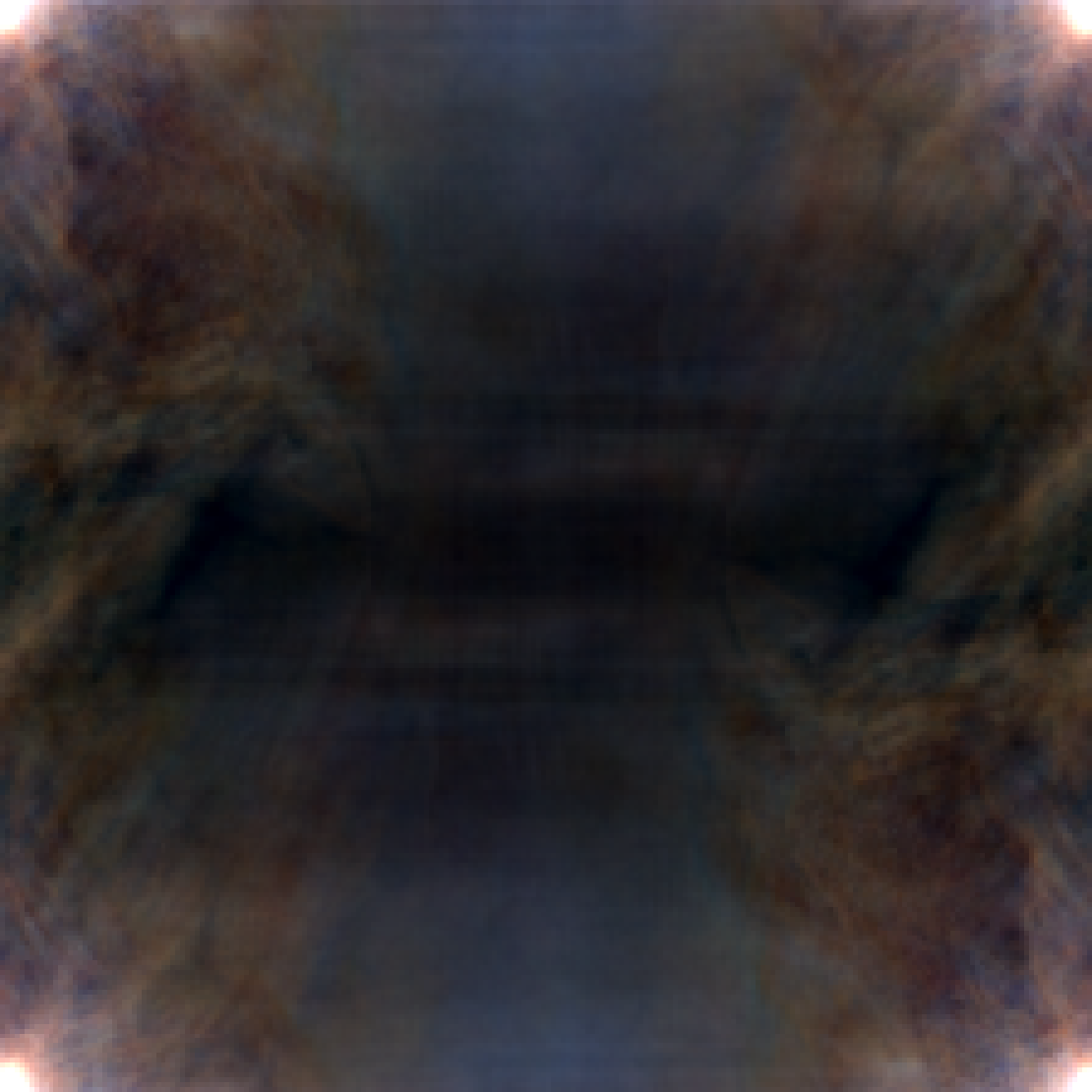}
		\includegraphics[width=0.19\linewidth]{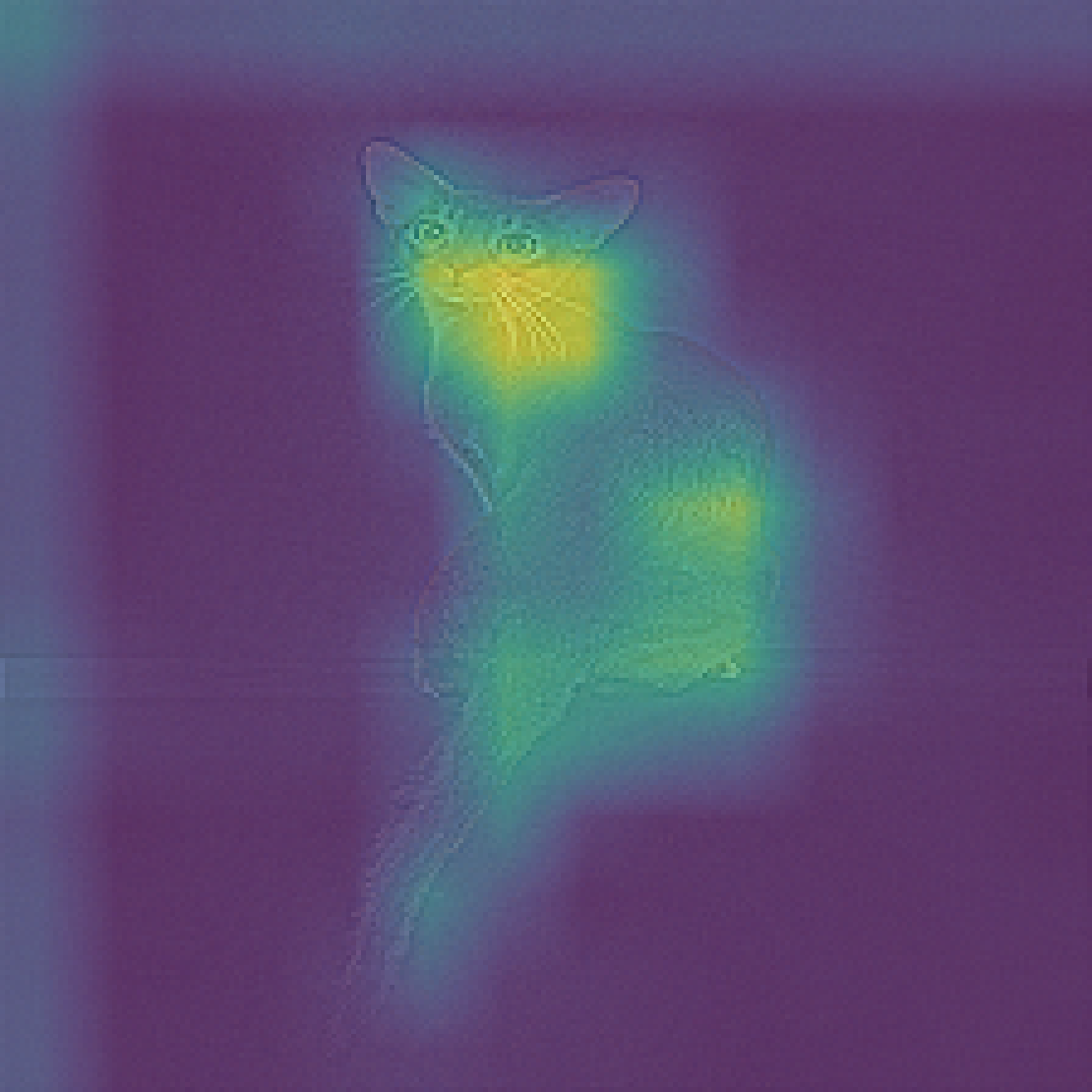}
		\includegraphics[width=0.19\linewidth]{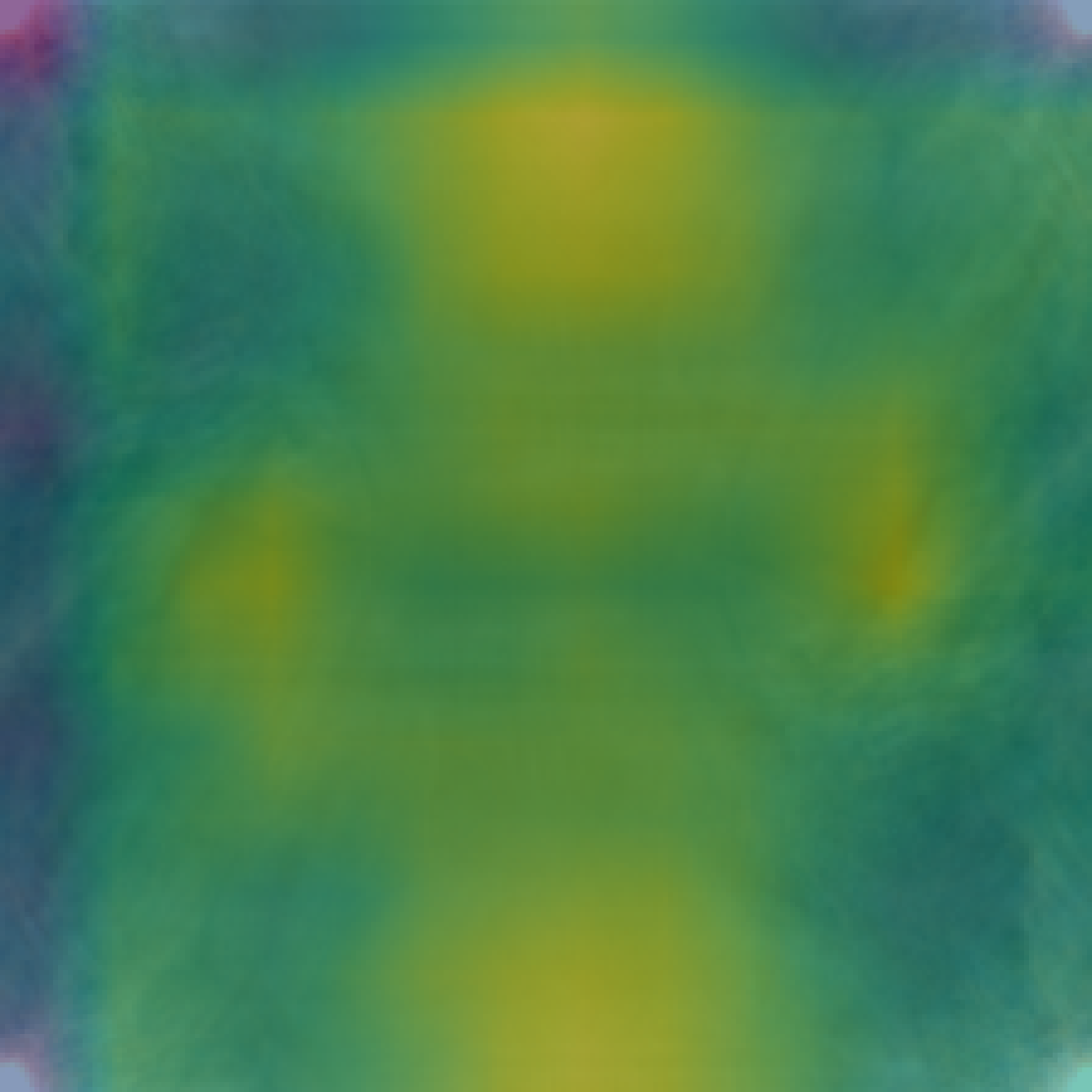}
	\end{subfigure}
	\caption{Illustration of image reconstruction from the Phase and Amplitude of the Fourier transform.}
	\label{fig:oxford_pets_fourier}
	\vspace{-0.5cm}
\end{figure}

The distinct roles of Fourier components are well-established in signal processing~\cite{phase_in_speech_pictures, Oppenheim1980TheIO}. As illustrated in Fig.~\ref{fig:oxford_pets_fourier}, the phase spectrum preserves semantic structure~\cite{piotrowski_demonstration_1982}, while the amplitude spectrum captures style~\cite{xu_fourier-based_2021, yang2020fdafourierdomainadaptation}. This property has been widely exploited in Domain Generalization (DG) and Unsupervised Domain Adaptation (UDA). Methods like FDA~\cite{yang2020fdafourierdomainadaptation} and FACT~\cite{xu_fourier-based_2021} typically employ Fourier Transform as a \textbf{data augmentation} technique. They synthesize new training samples by swapping the amplitude spectrum of a source image with that of a target image, thereby forcing the model to be invariant to amplitude shifts.

\textbf{Distinction of FARL:} While sharing the same theoretical foundation, FARL fundamentally differs from these DG methods in its methodology and application scope.
\begin{itemize}
	\item \textbf{Data-level vs. Representation-level:} Existing DG methods operate at the \textit{input data level} (augmentation). The downstream model still processes the augmented images holistically, leaving the internal feature entanglement unresolved. In contrast, FARL operates at the \textbf{representation level}. We design a dual-stream architecture that structurally enforces the separation of phase and amplitude features within the network's reasoning process.
	\item \textbf{Implicit Invariance vs. Explicit Disentanglement:} DG methods aim for implicit invariance by expanding the training distribution. FARL achieves explicit disentanglement via attention mechanisms, allowing the model to dynamically weigh the importance of structure versus style tokens for descriptive prompting in vision-language alignment. This makes FARL particularly suitable for few-shot VLM adaptation where large-scale retraining (required for effective augmentation) is not feasible.
\end{itemize}

\begin{figure*}[!t]             
	\centering
	\includegraphics[width=\textwidth]{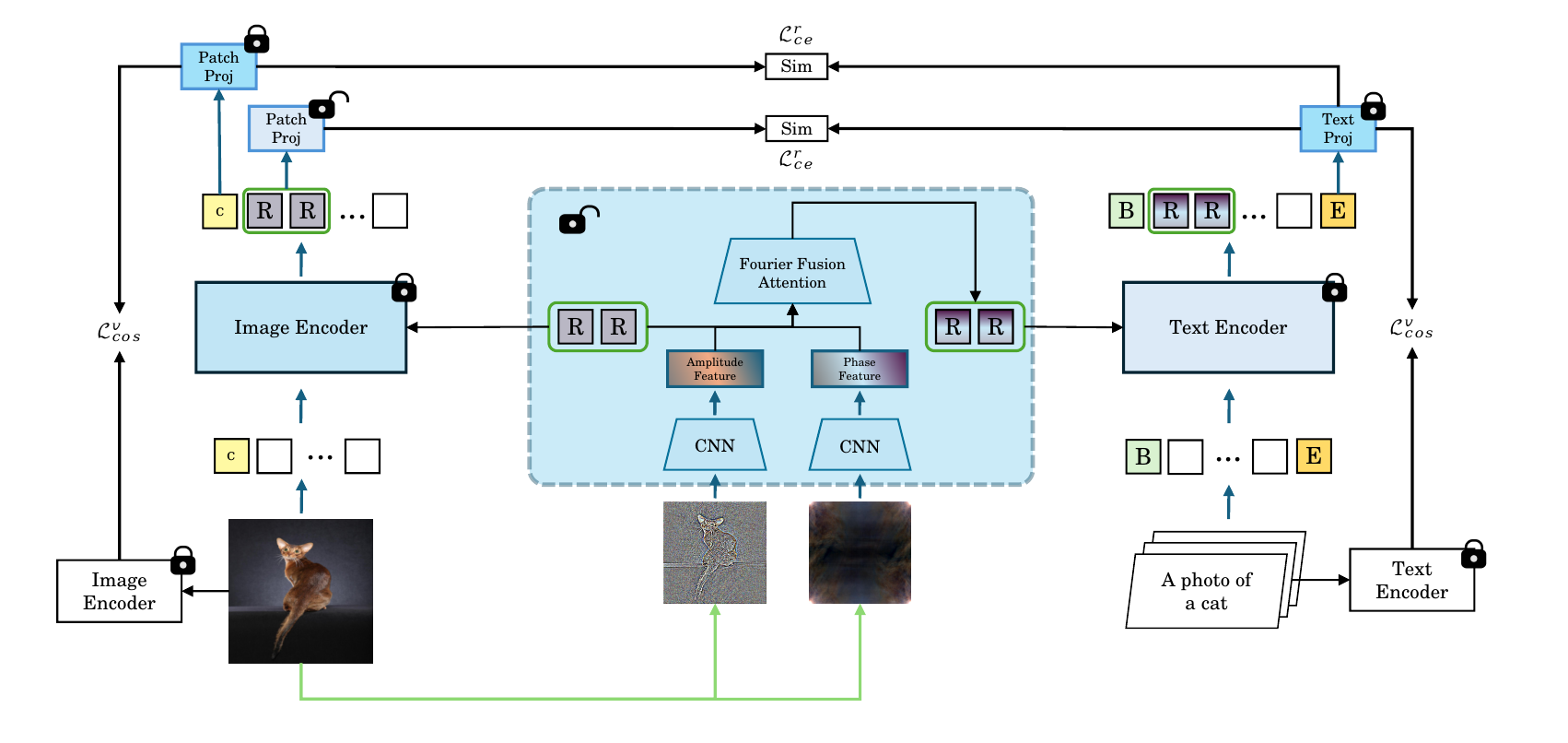}
	\caption{Overview of the FARL architecture. An image is decomposed into phase (structure) and amplitude (style) components. The Fourier Fusion Attention module (Fig. \ref{fig:cross_atn}) uses these disentangled features to enrich learnable representation tokens $R$. Following an asymmetric injection strategy, the fused tokens are injected into the Text Encoder, while the original $R$ tokens are injected into the Image Encoder. The model is optimized with a combination of cross-entropy $\mathcal{L}_{ce}$ and cosine regularization $\mathcal{L}_{cos}$ losses. Symbols: $c$: class token, $B/E$: text boundaries, $R$: representation tokens, $F$: projection layers.}
	\label{fig:pipeline}
	\vspace{-0.5cm}
\end{figure*}

\section{Method}
\label{sec:method}
Our proposed framework, \textbf{F}ourier-\textbf{A}ttentive \textbf{R}epresentation \textbf{L}earning (FARL), introduces a novel adaptation strategy for Vision-Language Models (VLMs) that directly addresses the challenge of feature entanglement by operating in the Fourier domain. The framework is designed to learn disentangled, interpretable, and generalizable representations from few-shot data. The overall pipeline is illustrated in Fig. \ref{fig:pipeline}.

\subsection{Preliminary: The CLIP Architecture}
\textbf{Image and Text Encoders:} The CLIP Image Encoder $\mathcal{V}$, and Text Encoder $\mathcal{T}$, are typically based on the Transformer architecture. For an input image $I$, $\mathcal{V}$ processes a sequence of patch embeddings $E_0$ prepended with a class token $c_0$. This sequence is passed through $L$ layers, and the final image feature $f_v$ is projected from the output class token $c_L$. Similarly, $\mathcal{T}$ processes a sequence of word embeddings $W_0$ framed by a Beginning-of-Text ($b_0$) and End-of-Text ($e_0$) token. This sequence is also passed through $L$ layers, and the final text feature $f_t$ is projected from the output EOT token $e_L$.

\begin{equation}
	[c_i, E_i] = \mathcal{V}_i([c_{i-1}, E_{i-1}]), \quad 1 \leq i \leq L
\end{equation}
\begin{equation}
	[b_i, W_i, e_i] = \mathcal{T}_i([b_{i-1}, W_{i-1}, e_{i-1}]), \quad 1 \leq i \leq L
\end{equation}

\textbf{Zero-Shot Classification:} CLIP performs zero-shot classification by computing the cosine similarity between an image feature $f_v$ and a set of text features $\{f_{t,c}\}_{c=1}^C$, generated from prompts for all $C$ classes. The prediction probability is given by a softmax over these similarities, enabling classification without task-specific training.

\subsection{FARL: A Fourier-Guided Approach}
The core hypothesis of FARL is that by explicitly separating an image's high-level, domain-invariant structure from its low-level, domain-specific style, we can guide a VLM to learn more robustly and avoid overfitting on superficial cues. Our framework achieves this through three key stages: (1) disentangling visual features via Fourier decomposition, (2) generating feature-rich representations using a dual-attention mechanism, and (3) asymmetrically adapting the VLM encoders with these representations.

\subsubsection{Fourier Decomposition and Feature Extraction:}
The initial stage of FARL is to decompose an input image $I \in \mathbb{R}^{B \times C \times H \times W}$ into two distinct components that isolate its structural and stylistic information. This process begins by applying a 2D Fast Fourier Transform:

\begin{equation}
	\mathcal{F}(I) = A \cdot e^{jP}
\end{equation}

where $A = |\mathcal{F}(I)|$ and $P = \angle \mathcal{F}(I)$ are the amplitude and phase spectra, respectively. Using the inverse FFT $\mathcal{F}^{-1}$, we reconstruct two separate images:

\textbf{Phase Image $I_{phase}$:} To isolate structural information, the phase image is reconstructed by preserving the original phase spectrum $P$ while setting the amplitude spectrum to unity. This component retains more high-level and domain-invariant features such as shapes and edges.

\begin{equation}
	I_{\text{phase}} = \text{Re}\left(\mathcal{F}^{-1}(1 \cdot e^{jP})\right)
\end{equation}

\textbf{Amplitude Image $I_{amp}$:} To isolate stylistic information, the amplitude image is reconstructed by preserving the original amplitude spectrum $A$ while setting the phase spectrum to zero. This component captures low-level and domain-specific statistics such as color, texture, and lighting.
\begin{equation}
	I_{\text{amp}} = \text{Re}\left(\mathcal{F}^{-1}(A \cdot e^{j \cdot 0})\right)
\end{equation}

After independent normalization, these two component images are passed through lightweight CNNs to extract sequences of patch tokens, yielding a structure-focused feature set $F_{phase}$ and a style-focused set $F_{amp}$.

These two images, after normalization, are passed through lightweight CNNs to produce sequences of patch tokens: a structure-focused feature set $F_{phase}$ and a style-focused set $F_{amp}$

\begin{figure*}[!t]             
	\centering
	\includegraphics[width=\textwidth]{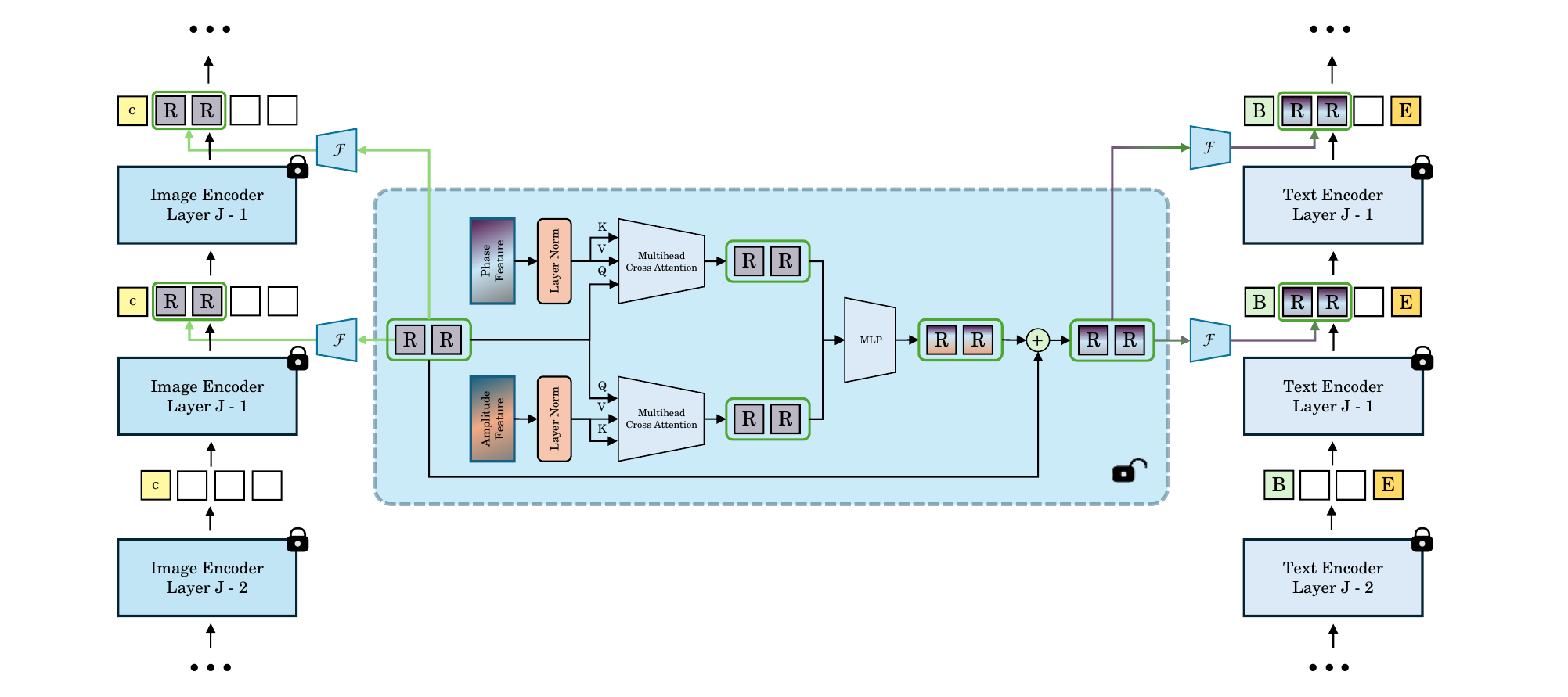}
	\caption{The Fourier Fusion Attention module. The module uses original representation tokens $R$ as Queries to attend to Phase and Amplitude Features as Keys/Values in parallel cross-attention blocks. The result are fused by an MLP and combined with the original $R$ via a residual connection to produce the final enriched tokens.}
	\label{fig:cross_atn}
\end{figure*}

\begin{figure}[h]
	\centering
	\begin{subfigure}{\linewidth}
		\centering
		\raisebox{0.2cm}{\begin{turn}{90}\tiny{OxfordFlowers}\end{turn}}
		\includegraphics[width=0.18\linewidth]{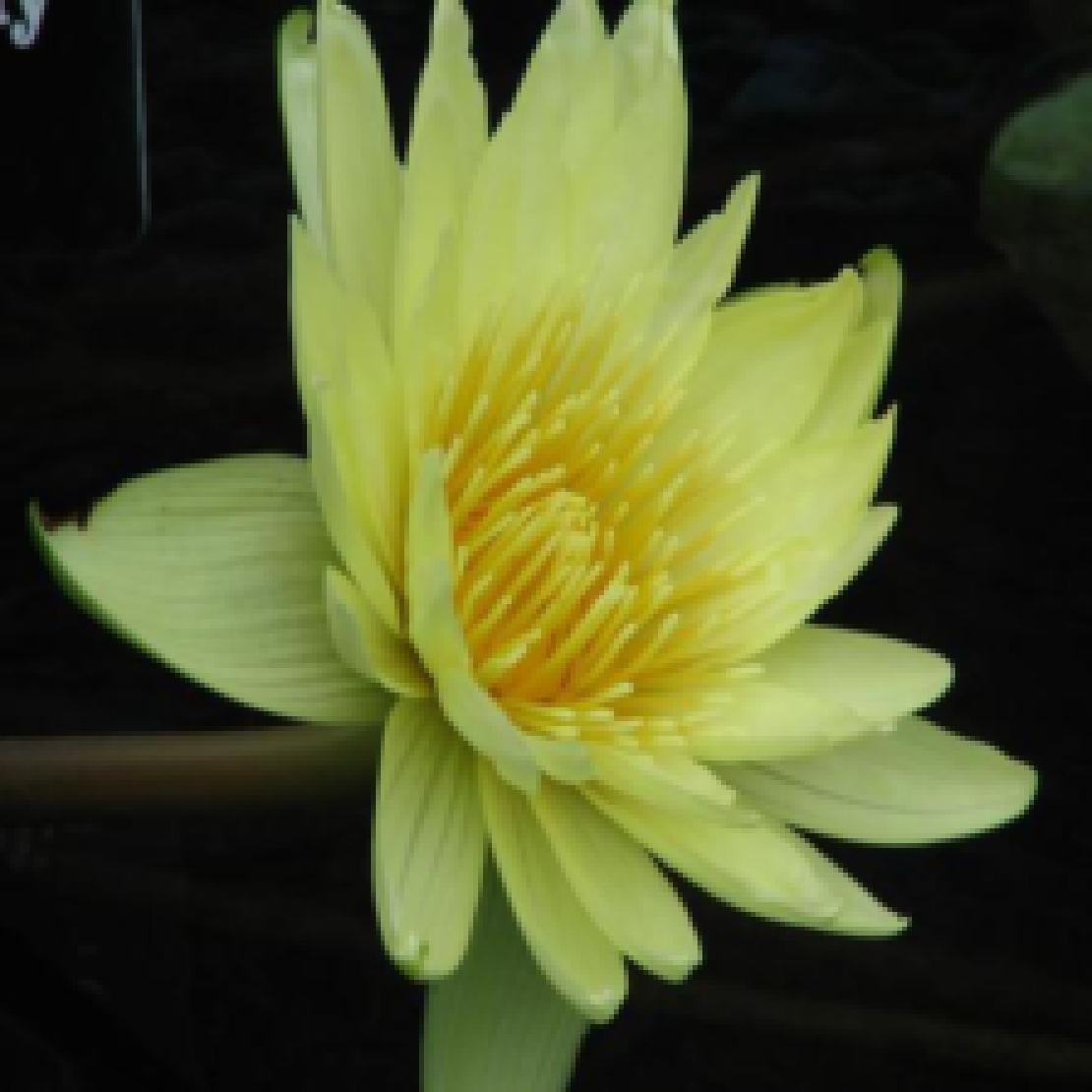}
		\includegraphics[width=0.18\linewidth]{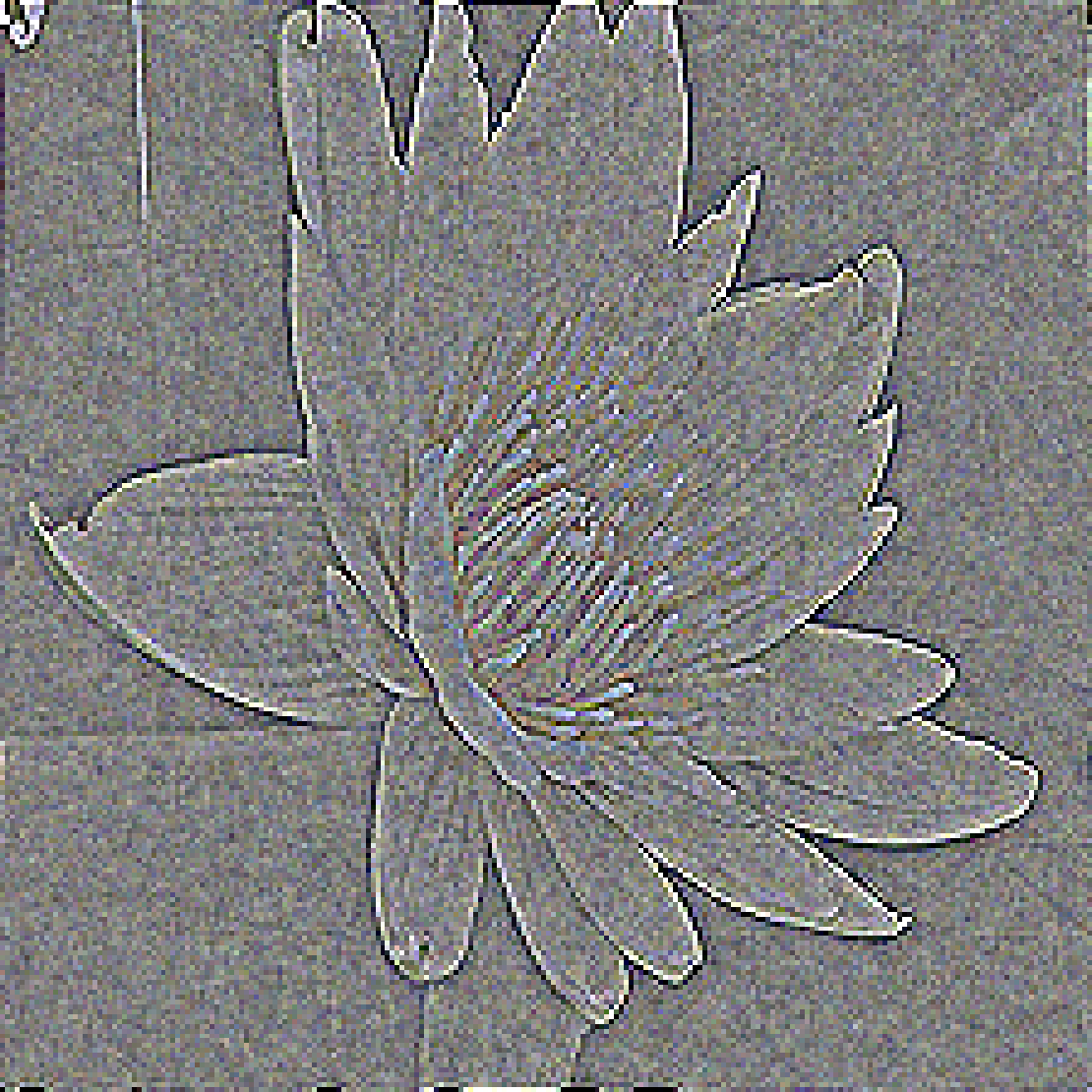}
		\includegraphics[width=0.18\linewidth]{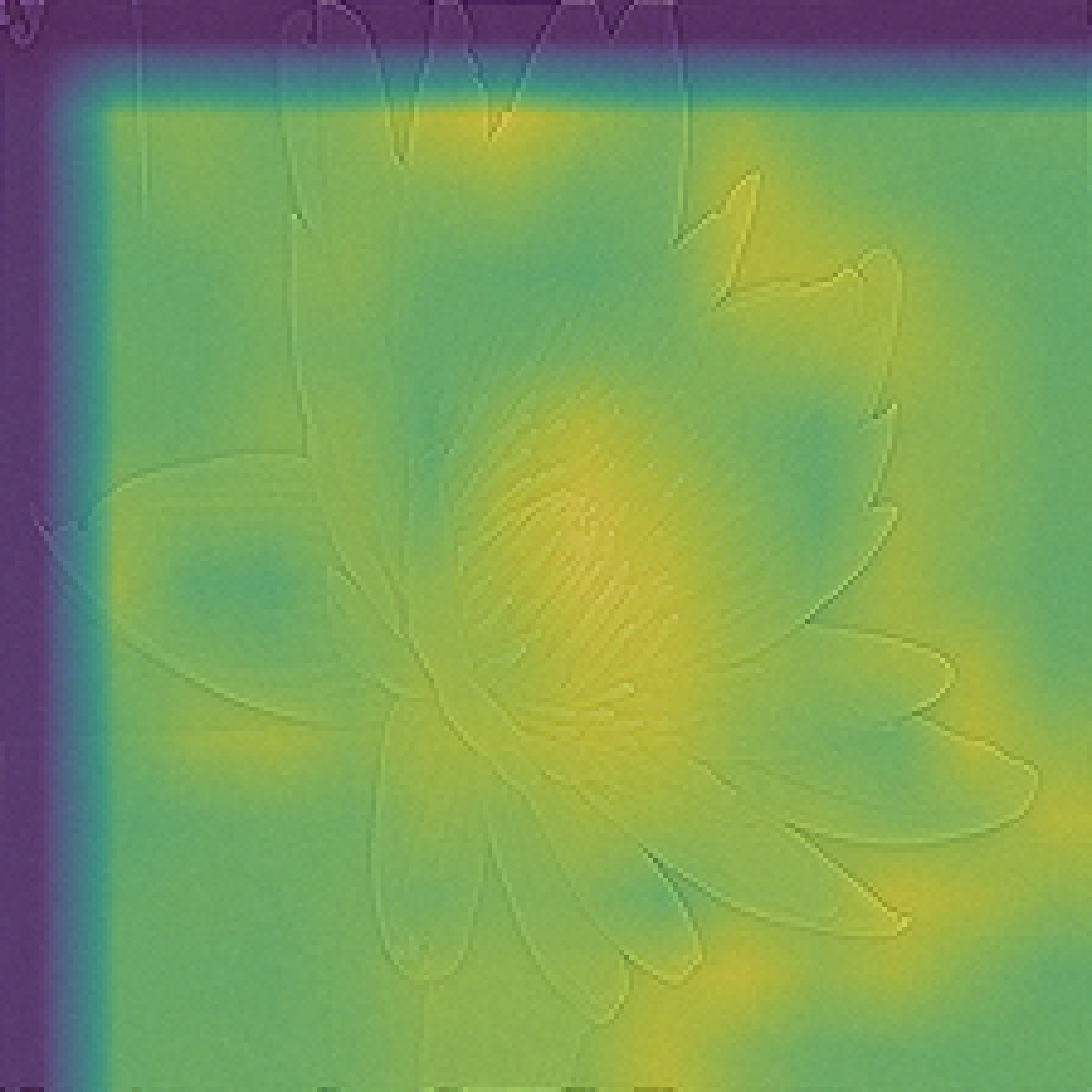} 
		\includegraphics[width=0.18\linewidth]{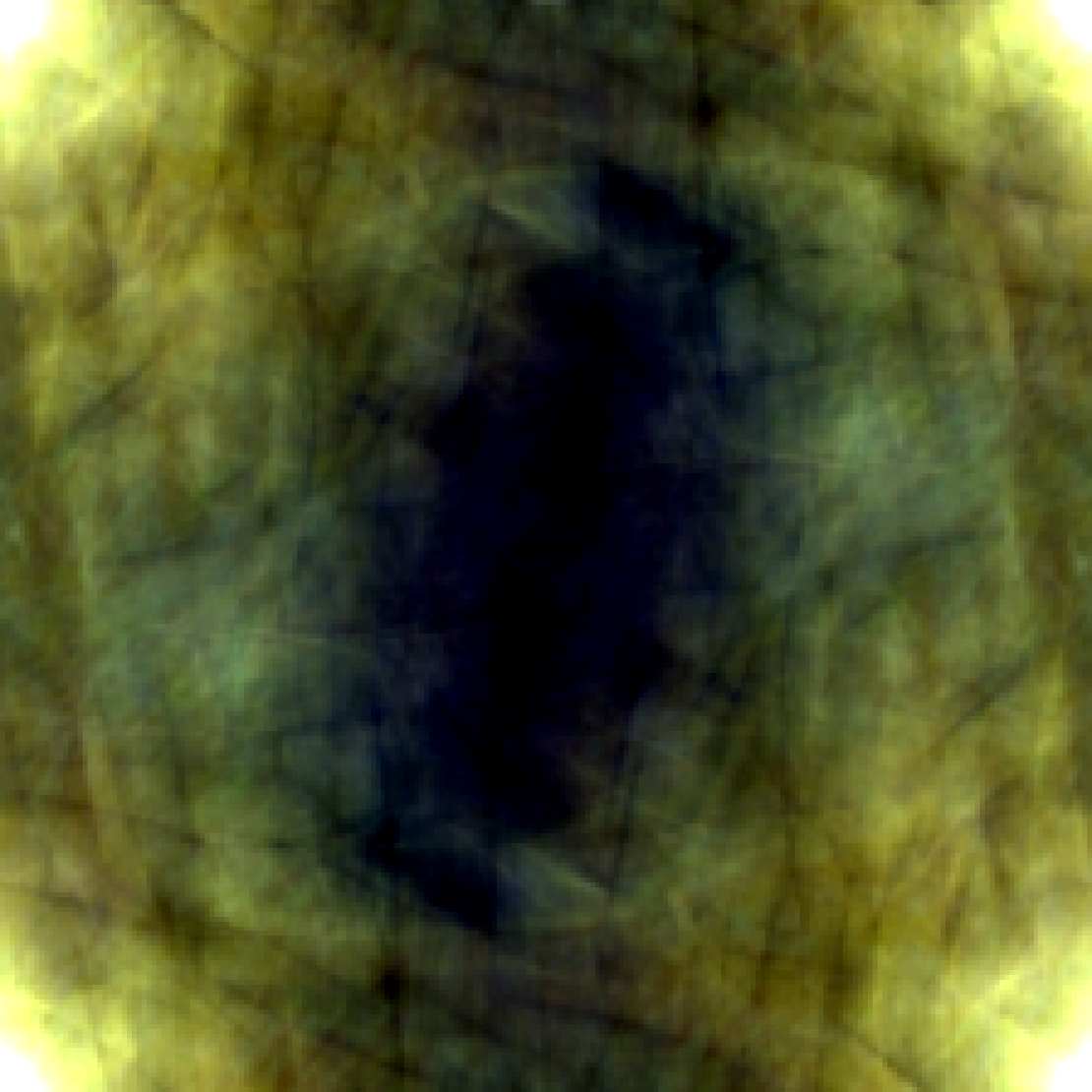}       \includegraphics[width=0.18\linewidth]{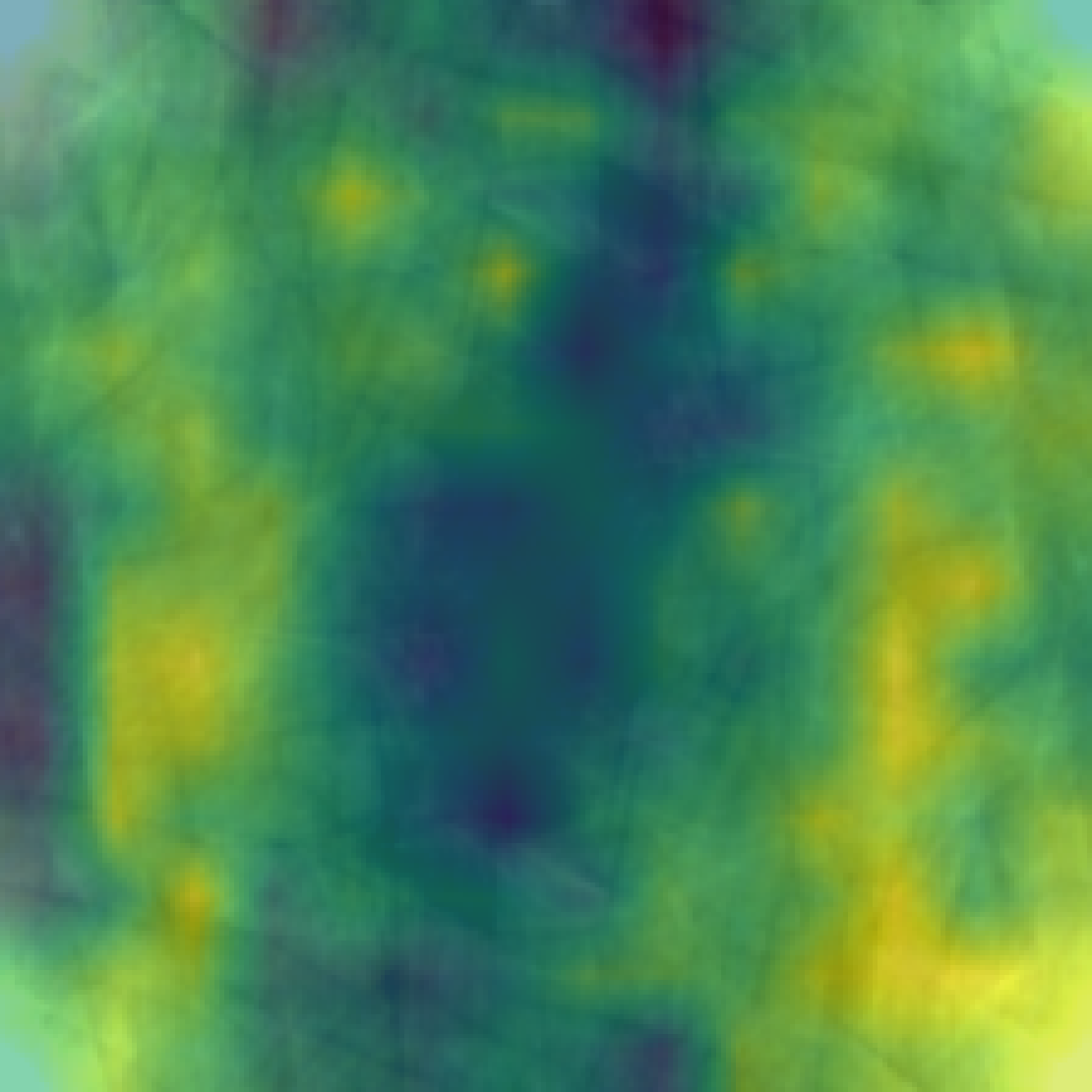}
	\end{subfigure}
	
	\begin{subfigure}{\linewidth}
		\centering
		\raisebox{0.2cm}{\begin{turn}{90}\tiny{FGVCAircraft}\end{turn}}
		\includegraphics[width=0.18\linewidth]{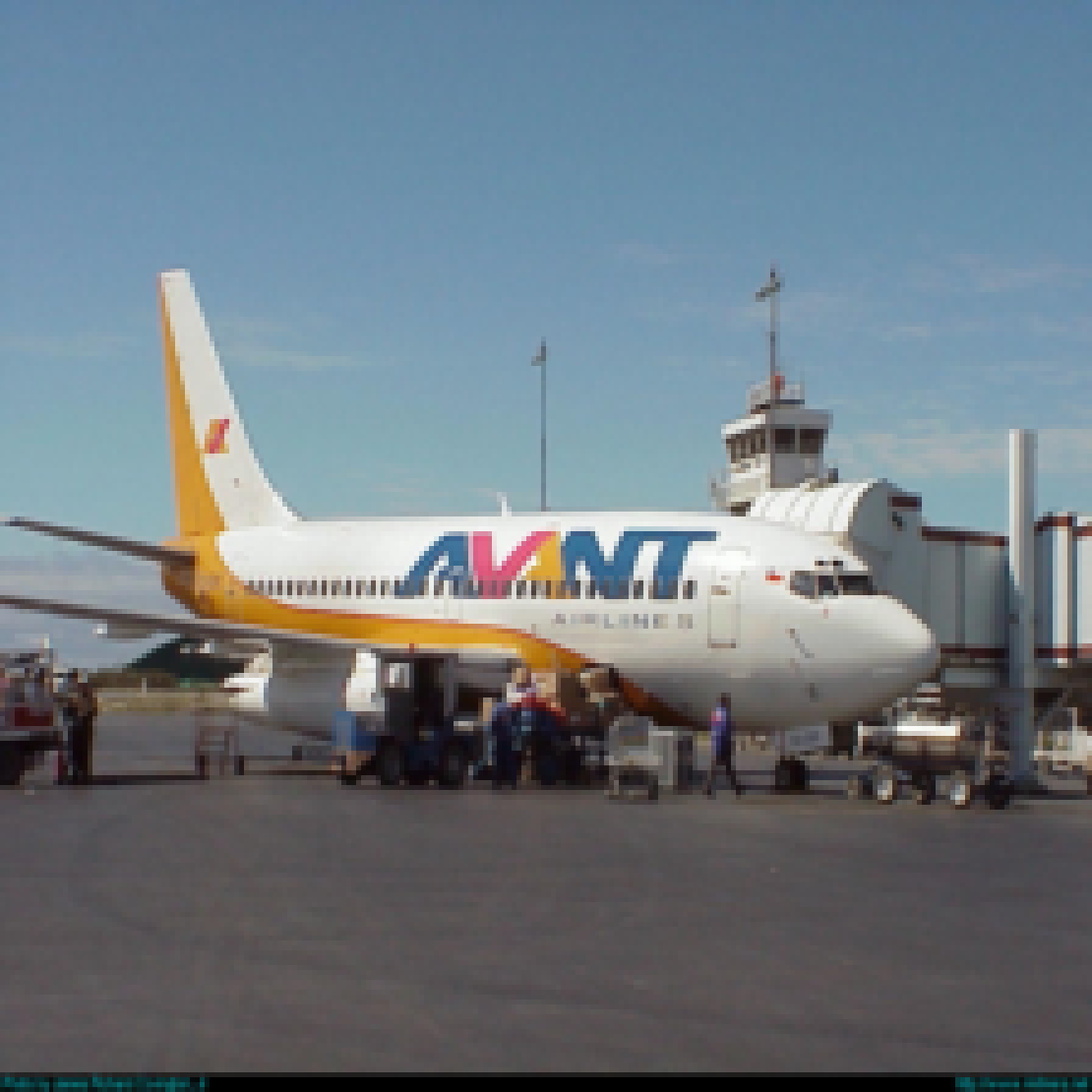}
		\includegraphics[width=0.18\linewidth]{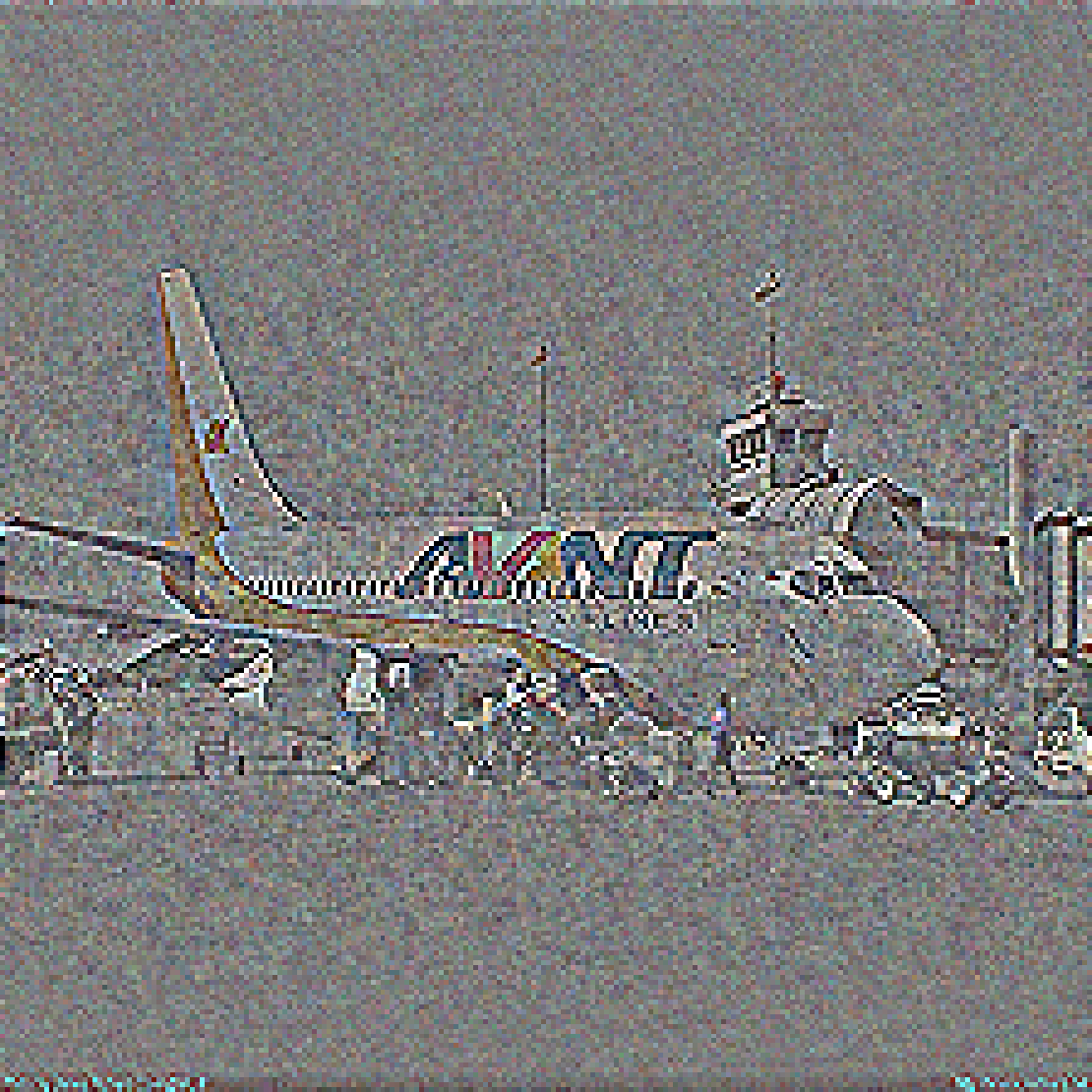}
		\includegraphics[width=0.18\linewidth]{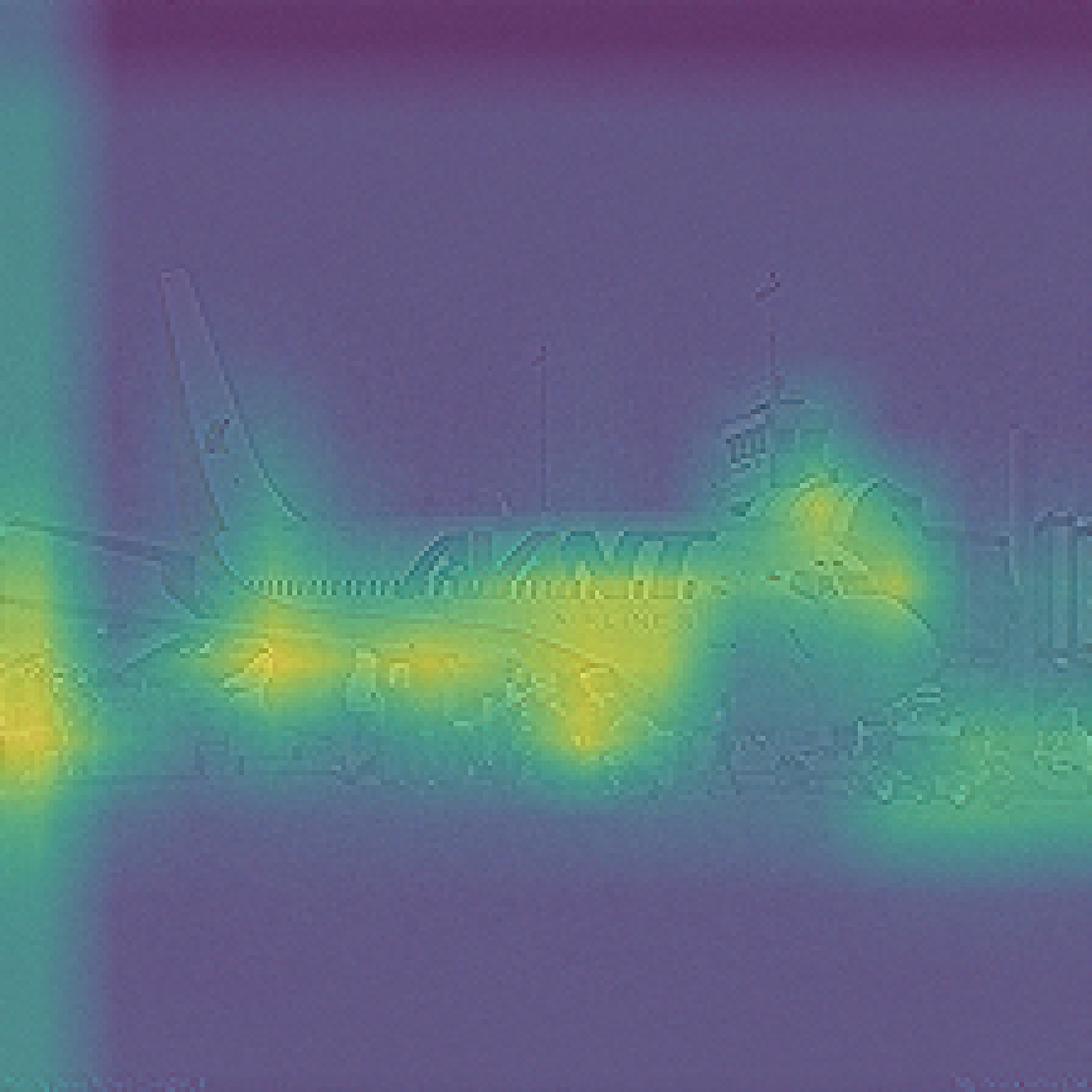}
		\includegraphics[width=0.18\linewidth]{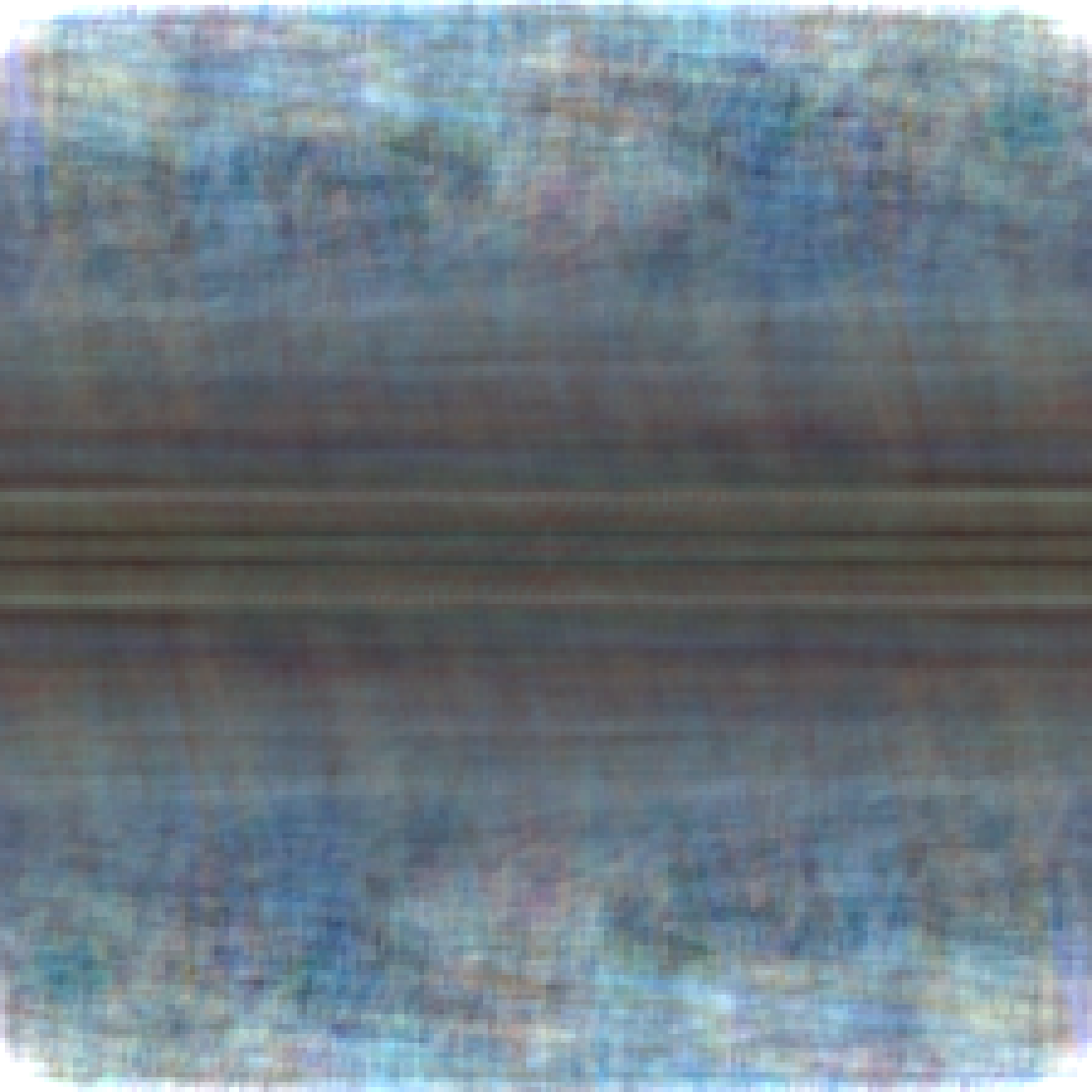}
		\includegraphics[width=0.18\linewidth]{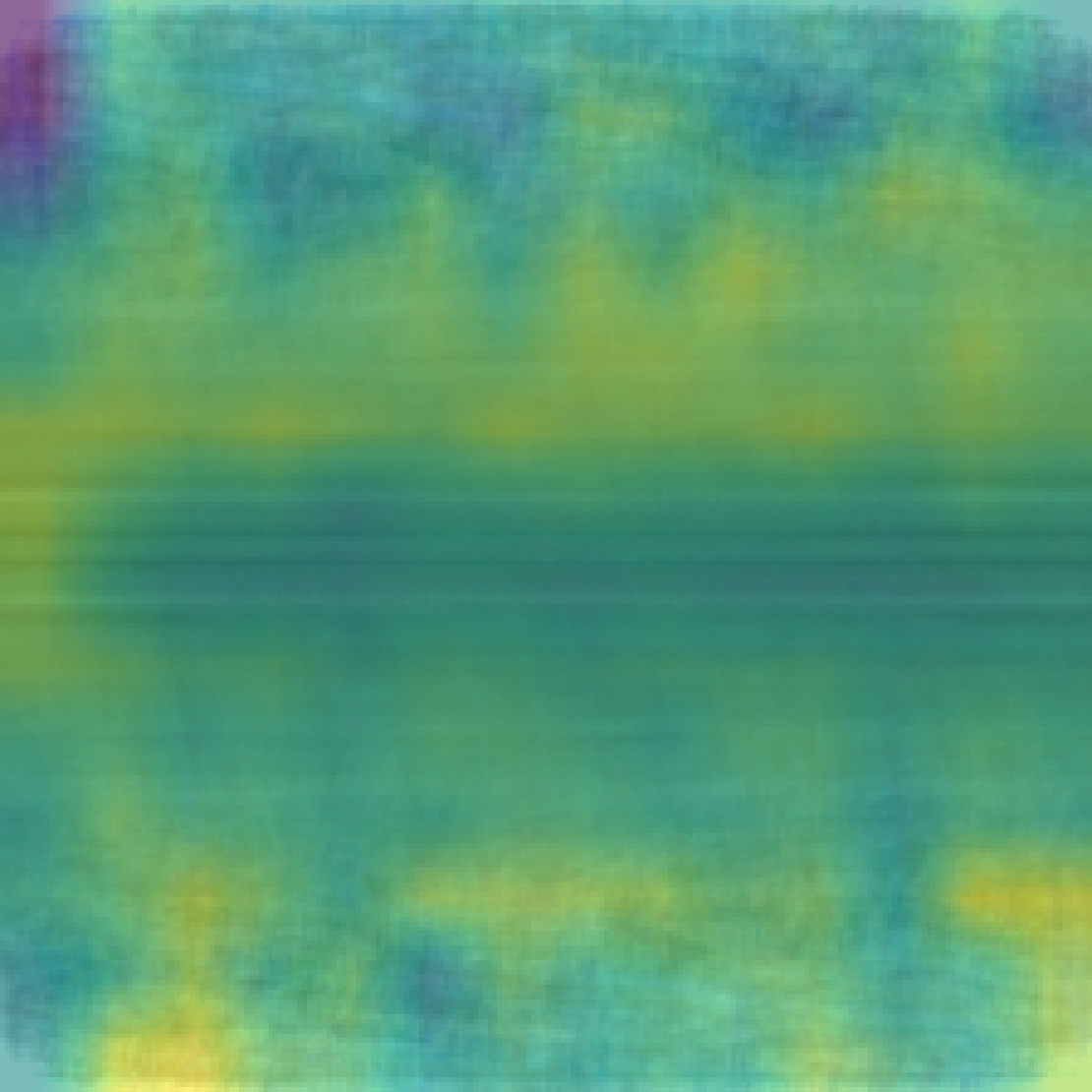}
	\end{subfigure}
	
	\begin{subfigure}{\linewidth}
		\centering
		{\begin{turn}{90}\tiny{DescribableTextures}\end{turn}}
		\includegraphics[width=0.18\linewidth]{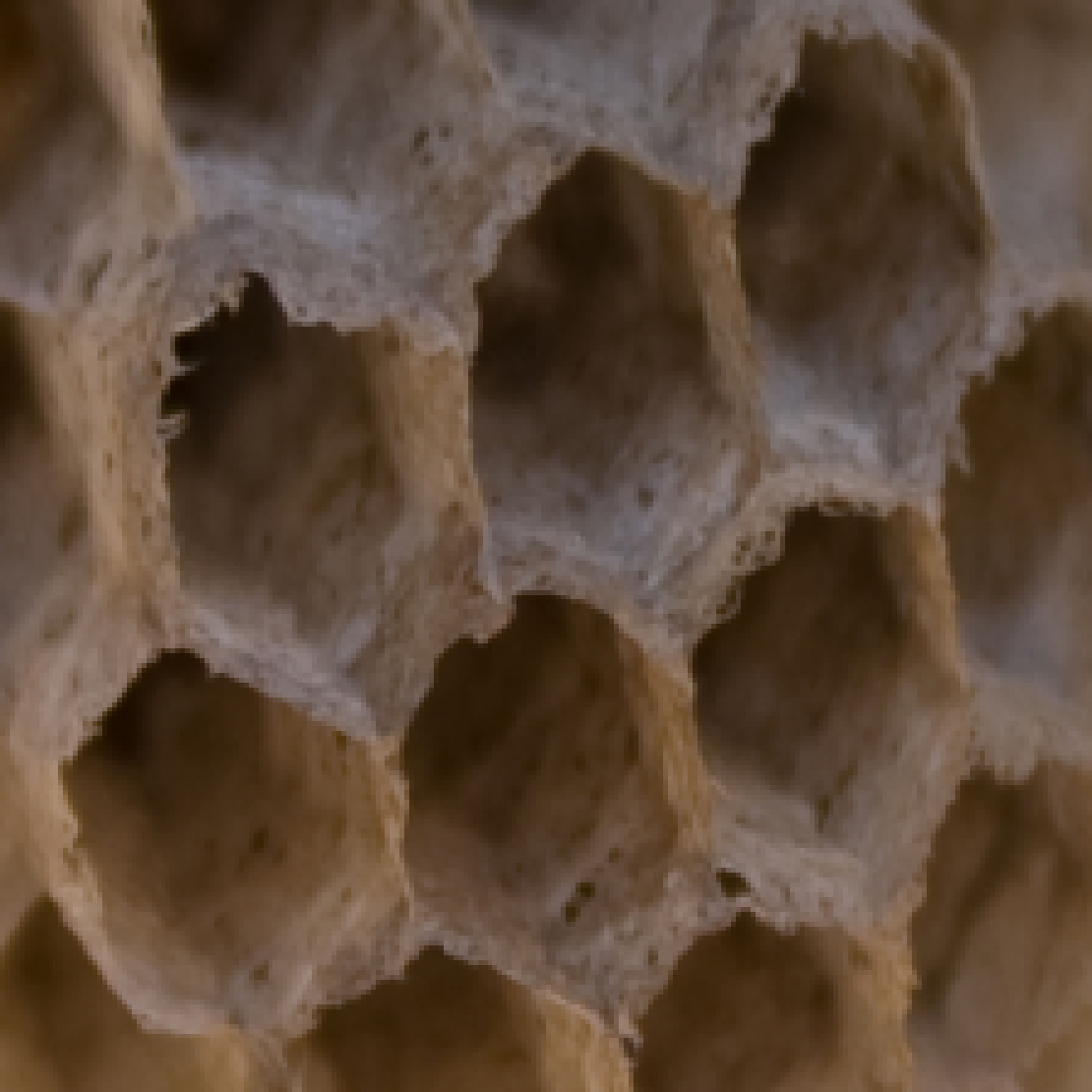}
		\includegraphics[width=0.18\linewidth]{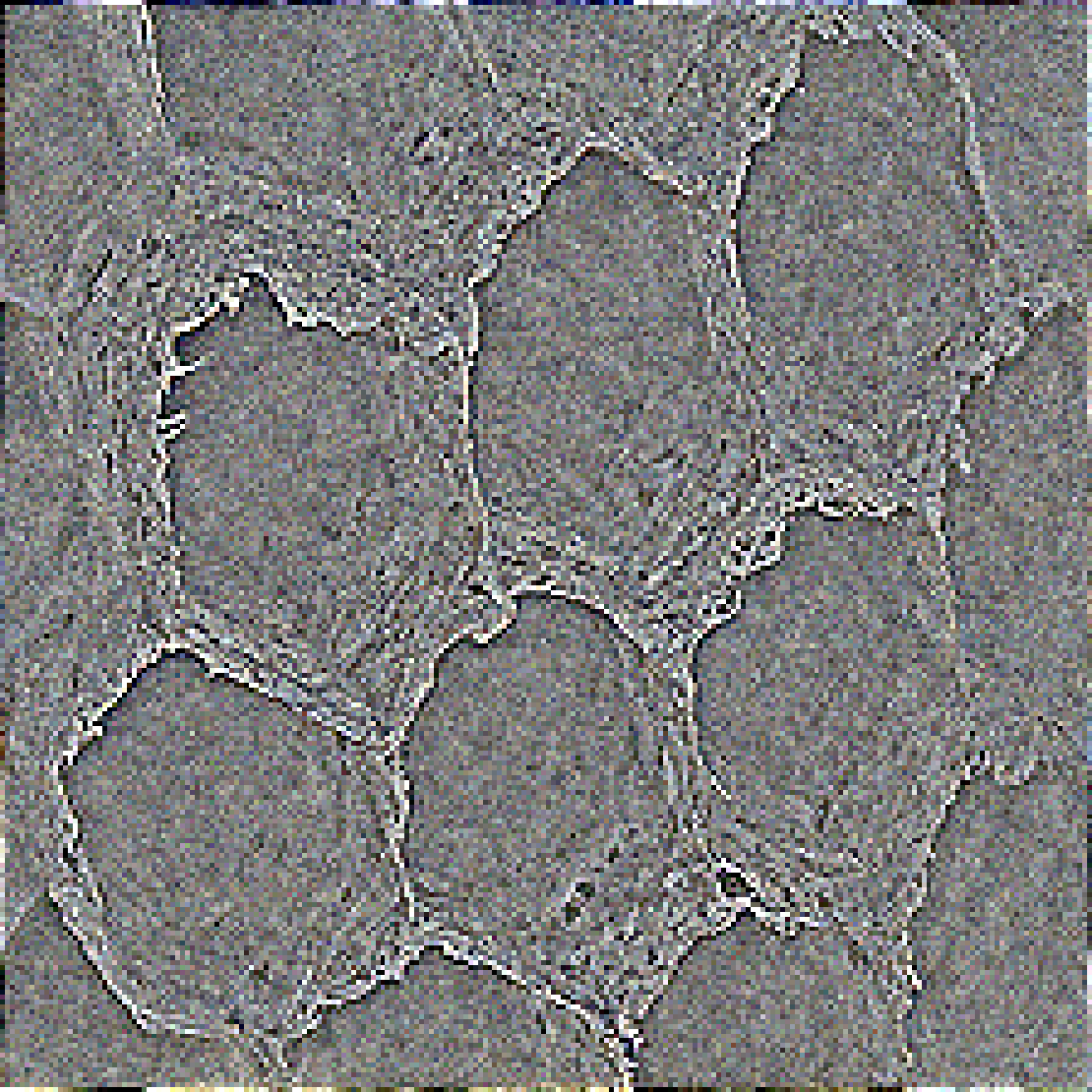}
		\includegraphics[width=0.18\linewidth]{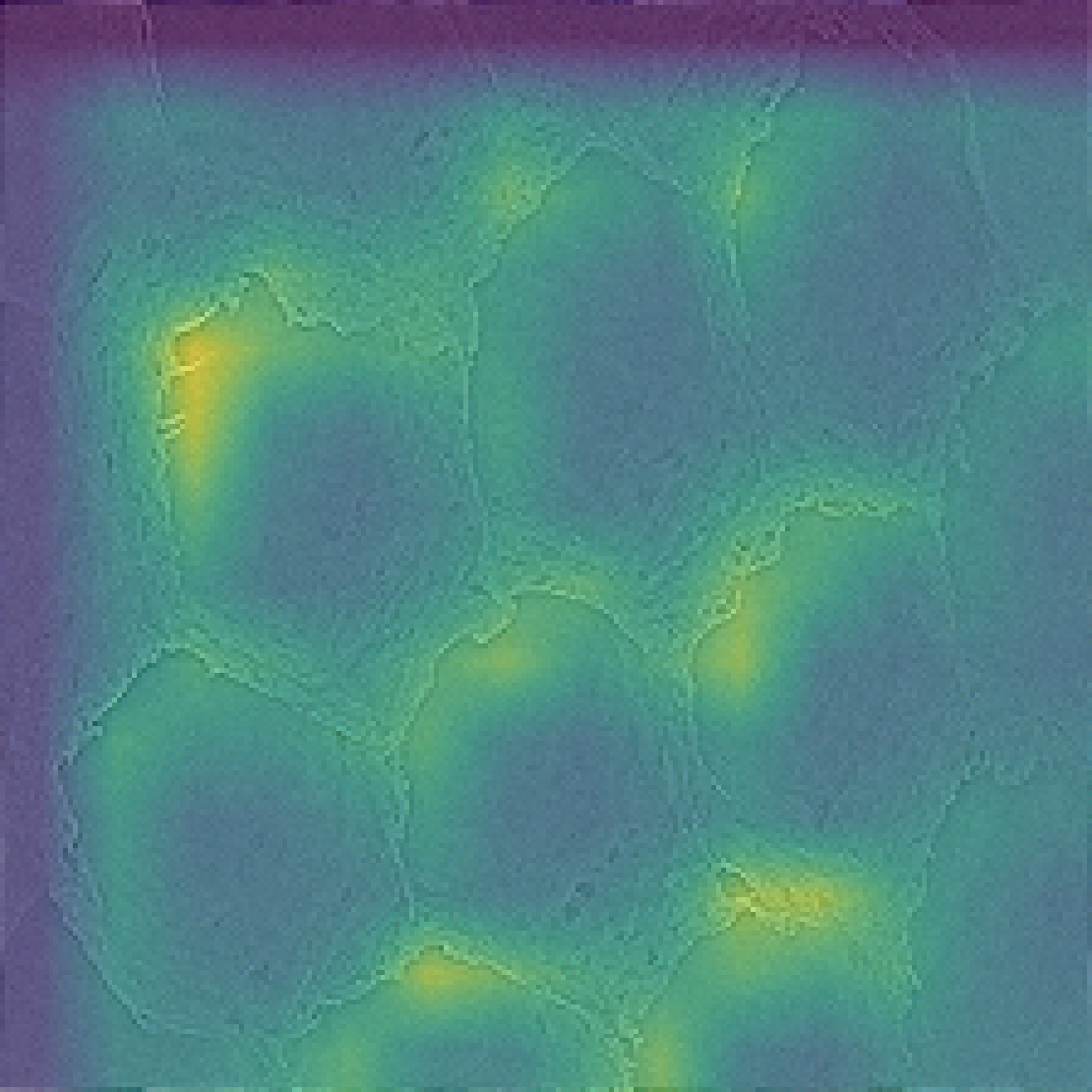}
		\includegraphics[width=0.18\linewidth]{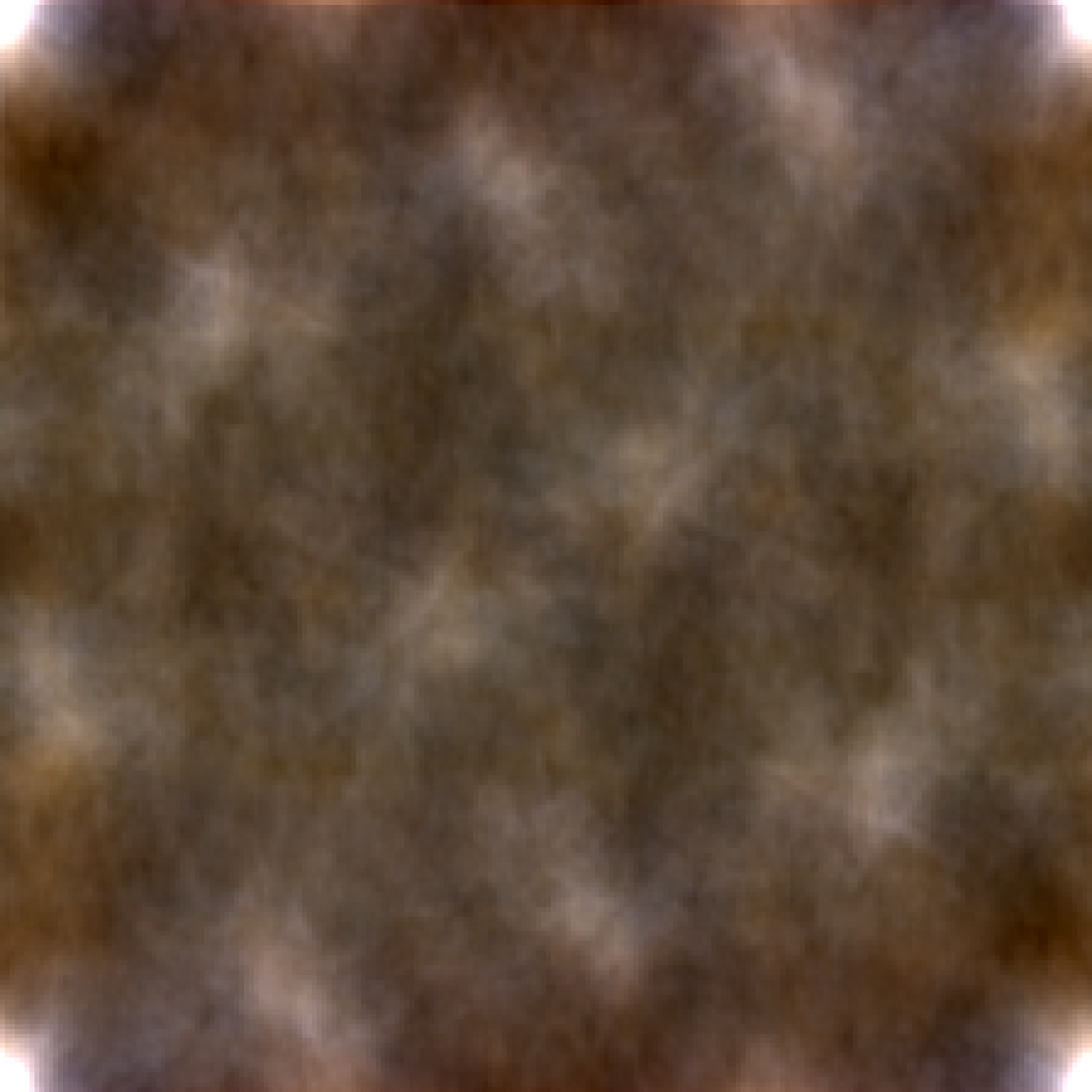}
		\includegraphics[width=0.18\linewidth]{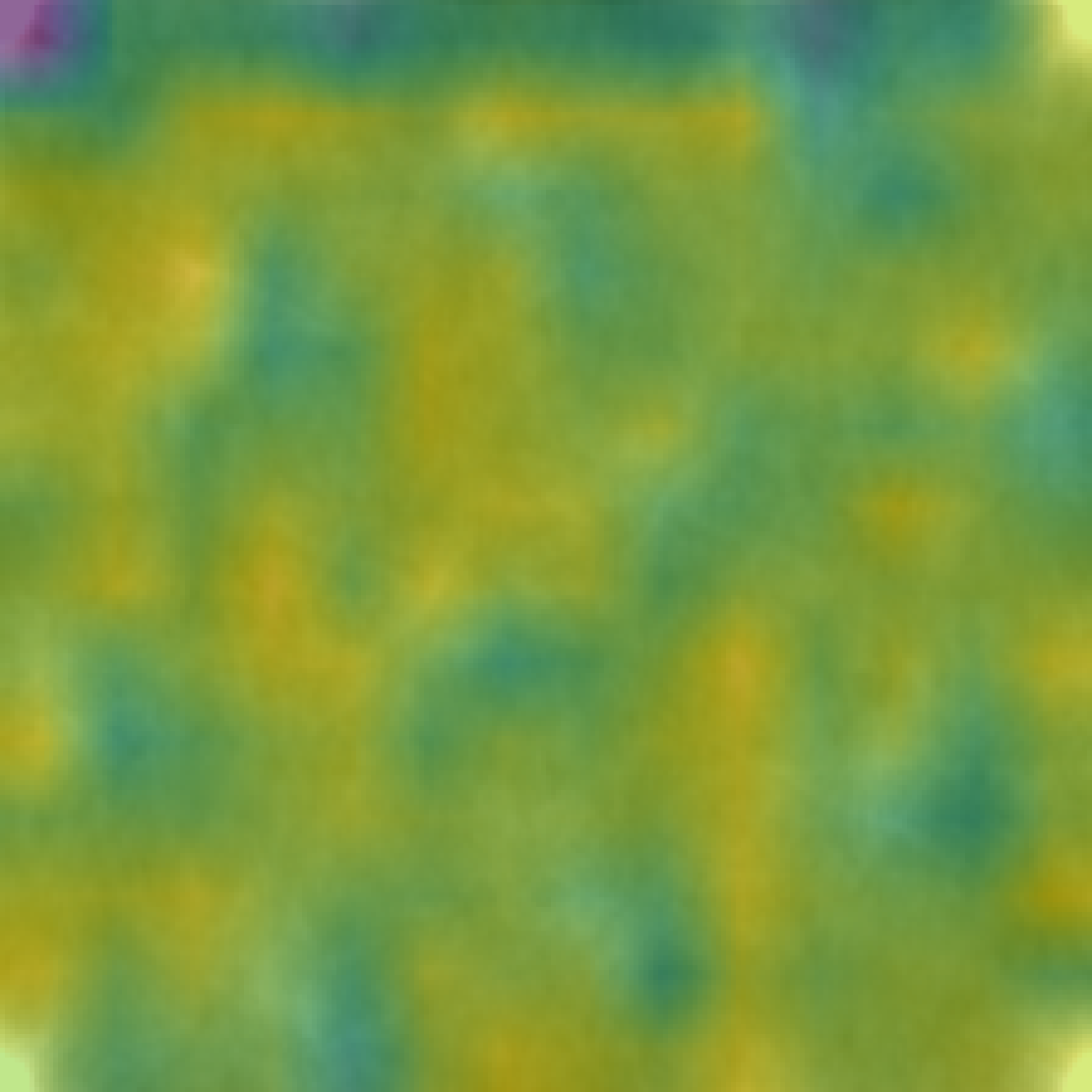}
	\end{subfigure}
	
	\caption{Visualization of Fourier decomposition and the dual-attention mechanism across diverse datasets. From left to right, each row displays: (1) the original image, (2) the phase-only reconstruction, (3) the attention map from the phase stream, (4) the amplitude-only reconstruction, and (5) the attention map from the amplitude stream.}
	\label{fig:fourier_and_attn}
	\vspace{-0.5cm}
\end{figure}

\subsubsection{Fourier-Attentive Representation Fusion:}
To leverage these disentangled features, FARL introduces a set of $K$ learnable, modality-agnostic conceptual tokens, $R \in \mathcal{R}^{K \times D_{rep}}$. These tokens act as trainable queries to probe the two visual streams via a dual cross-attention mechanism, as shown in Fig. \ref{fig:pipeline}, \ref{fig:cross_atn}.

The mechanism generates two specialized sets of tokens: structure-aware $R'_{phase}$ and style-aware $R'_{amp}$ tokens:

\begin{equation}
	R'_{\text{phase}} = \text{CrossAttn}_{\text{phase}}(Q=R; \; K, V=F_{\text{phase}})
\end{equation}
\begin{equation}
	R'_{\text{amp}} = \text{CrossAttn}_{\text{amp}}(Q=R; \; K, V=F_{\text{amp}})
\end{equation}

These two sets are then fused by a fusion MLP $\mathcal{M}_{fuse}$ and integrated back into the original tokens via a weighted residual connection to produce the final, enriched representation tokens, $R_{fused}$.

\begin{equation}
	R_{fusion} = \mathcal{M}_{\text{fuse}}(\text{Concat}(R'_{\text{phase}}, R'_{\text{amp}}))
\end{equation}
\begin{equation}
	R_{fused} = R + R_{\text{fusion}}
\end{equation}

The resulting $R_{fused}$ tokens now serve as a rich, disentangled semantic foundation for adapting the VLM.

\subsubsection{Asymmetric Injection and Decoupled Inference:}
FARL employs an asymmetric adaptation strategy to inject the learned representations into the higher layers (from layer $J$ onwards) of the VLM's encoders.

\textbf{Text Encoder:} The enriched, fused tokens $R_{fused}$, which carry the disentangled structural and stylistic information, are projected and subsequently injected into the higher layers of the Text Encoder. For each target layer $j$ (where $J \leq j \leq L$), a dedicated projection function $P_j^t$ maps the tokens:

\begin{equation}
	R^t_j = \mathcal{P}^t_j(R_{\text{fused}})
\end{equation}
\begin{equation}
	[b_j, R^t_j, W_j, e_j] = \mathcal{T}_j([b_{j-1}, R^t_{j-1}, W_{j-1}, e_{j-1}])
\end{equation}

The final text feature for a class $c$, $f_{t,c}$, is thus conditioned on the disentangled visual information, forming a richer, instance-specific representation.

\textbf{Image Encoder:} In contrast, the original, more general representation tokens $R$ are projected and injected into the image encoder\cite{mmrl}. This process yields two distinct visual features: Class Feature $f_v$, derived from the original class token $c$ path and Representation Feature $f_r$, derived from the injected representation tokens. For each target layer $j$ (where $J \leq j \leq L$):

\begin{equation}
	R^v_i = \mathcal{P}^v_i(R)
\end{equation}
\begin{equation}
	[c_i, R^v_i, E_i] = \mathcal{V}_i([c_{i-1}, R^v_{i-1}, E_{i-1}])
\end{equation}

\textbf{Decoupled Learning Strategy:} The two visual features, $f_v$ and $f_r$, are leveraged differently for training and inference.

The training loss optimizes both features simultaneously, while a regularization term ensures the class feature path does not deviate from the original CLIP space\cite{mmrl}\cite{clip}:

\begin{equation}
	\mathcal{L} = \alpha \mathcal{L}_{ce}(f_c) + (1-\alpha) \mathcal{L}_{ce}(f_r) + \lambda(\mathcal{L}_{\text{cos}}^v + \mathcal{L}_{\text{cos}}^t)
\end{equation}

A key strategy for robust generalization is the decoupled use of features during inference\cite{mmrl}:

\begin{itemize}
	\item For \textbf{Base Classes} (seen during training): the final prediction combines the logits of both the class feature $f_v$ and the representation feature $f_r$, leveraging both general and task-specific knowledge.
	\item  For \textbf{Novel Classes} (unseen during training): the prediction relies only on the more general class feature $f_c$, which retains more of the pre-trained model's robust, zero-shot capabilities.
\end{itemize}

\subsection{Why Asymmetric Injection Matters?}
\label{sec:why_asymmetric}

A critical design choice in FARL is the asymmetric treatment of the two encoders: injecting enriched $R_{fused}$ into the text side while keeping the image side conditioned on generic $R$. We justify this design through the lens of \textbf{Semantic Abstraction} versus \textbf{Visual Regularization}.

\textbf{Text Side: Instance-Specific Semantic Abstraction:}
The role of the text encoder in VLM adaptation is to construct a semantic target (a classifier weight) that matches the visual input. In few-shot scenarios, a generic prompt like "a photo of a dog" is often too broad. By injecting $R_{fused}$, which carries explicit information about the instance's structure (phase) and style (amplitude), into the text encoder, we effectively convert the prompt into an instance-specific description (e.g., implicitly representing "a photo of a \textit{fluffy, white} dog"). This allows the text encoder to dynamically adjust its semantic focus to match the spectral properties of the input image, bridging the modality gap.

\textbf{Image Side: Generic Visual Regularization:}
Conversely, injecting the highly specific $R_{fused}$ into the image encoder would be detrimental. The visual backbone of CLIP is already powerful; forcing it to process tokens heavily biased towards the training instance's spectral statistics (especially amplitude/style) risks \textbf{overfitting}. By injecting the generic, randomly initialized tokens $R$ into the image encoder, we treat them as a regularization mechanism. They act as static "anchors" that preserve the pre-trained, general-purpose nature of the visual features, preventing the image encoder from drifting into a domain-specific subspace defined by the few-shot support set.

\textbf{Comparison with Alternatives:}
Experiments (Tabl. \ref{tab:variant}) confirm this hypothesis. \textit{Symmetric injection} (putting $R_{fused}$ into both) leads to performance degradation, likely because the image encoder overfits the style cues. Our asymmetric design strikes the optimal balance: rich semantic guidance for the text, and robust regularization for the image.
\begin{table*}
	\centering
	\resizebox{\textwidth}{!}{
		\begin{tabular}{lc|ccccccccccc|c}
			\toprule
			Datasets&  Sets&   CLIP&CoOp&  CoCoOp
			&    KgCoOp
			&PLOT
			&MaPLe&    ProVP&MetaPrompt&TCP&  MMA&  MMRL &\textbf{FARL}\\
			&  &   \scriptsize{(ICML2021)}&\scriptsize{(IJCV22)}&  \scriptsize{(CVPR22)}&    \scriptsize{(CVPR23)}&\scriptsize{(ICLR23)}&\scriptsize{(CVPR23)}&    \scriptsize{(IJCV2024)}&\scriptsize{(TIP2024)}&\scriptsize{(CVPR2024)}&  \scriptsize{(CVPR2024)}&   \scriptsize{(CVPR2025)}&\scriptsize{(our)}\\
			\midrule
			&  Base&   69.34&82.38&  80.47&    80.73&83.98&82.28&    85.20&83.66&84.13&  83.20&   85.66&\textbf{86.11}\\
			Average&  New&   74.22&67.96&  71.69&    73.61&71.72&75.14&    73.23&75.48&75.36&  76.94&   76.19&\textbf{77.49}\\
			&  HM&   71.59&73.89&  75.44&    76.71&76.76&78.27&    78.38&79.08&79.26&  79.72&   80.65&\textbf{81.57}\\
			\midrule
			&  Base&   72.43&76.46&  75.98&    75.83&77.30&76.66&    75.82&77.52&77.27&  77.31&   77.80&\textbf{78.03}\\
			ImageNet&  New&   68.14&66.31&  70.43&    69.96&69.87&70.54&    69.21&70.83&69.87&  71.23&   71.13&\textbf{71.33}\\
			& HM&  70.22&71.02& 73.10& 72.78& 73.40& 73.47& 72.36& 74.02& 73.38& 74.02& 74.37&\textbf{74.53}\\
			\midrule
			& Base&  96.84&97.80& 97.96& 97.72& 98.53& 97.74& 98.92& 98.13& 98.23& 98.40& 98.83&\textbf{99.23}\\
			Caltech101& New&  94.00&93.27& 93.81& 94.39& 92.80& 94.36& 94.21& 94.58& 94.67& 94.00& 94.50&\textbf{94.93}\\
			& HM&  95.40&95.48& 95.84& 96.03& 95.58& 96.02& 96.51& 96.32& 96.42& 96.15& 96.62&\textbf{97.03}\\
			\midrule
			& Base&  91.17&94.47& 95.20& 94.65& 94.50& 95.43& 95.87& 95.53& 94.67& 95.40& 95.87&\textbf{96.10}\\
			OxfordPets& New&  97.26&96.00& 97.69& 97.76& 96.83& 97.76& 97.65& 97.00& 97.20& \textbf{98.07}& 97.27&\textcolor{blue}{\textit{97.63}}\\
			& HM&  94.12&95.23& 96.43& 96.18& 95.65& 96.58& 96.75& 96.26& 95.92& 96.72& 96.56&\textbf{96.86}\\
			\midrule
			& Base& 63.37& 75.67& 70.49& 71.76& 79.07& 72.94& 80.43& 76.34& 80.80& 78.50& \textbf{81.50}&81.43\\
			StanfordCars& Novel& 74.89& 67.53& 73.59& 75.04& 74.80& 74.00& 67.96& 75.01& 74.13& 73.10& 74.90&\textbf{75.10}\\
			& HM& 68.65& 71.37& 72.01& 73.36& 76.88& 73.47& 73.67& 75.48& 77.32& 75.70& 78.06&\textbf{78.14}\\
			\midrule
			& Base&  72.08&97.27& 94.87& 95.00& 97.93& 95.92& 98.42& 97.66& 97.73& 97.77& 98.87&\textbf{98.9}\\
			Flowers& New&  \textbf{77.80}&67.13& 71.75& 74.73& 73.53& 72.46& 72.06& 74.49& 75.57& 75.93& 77.23&77.1\\
			& HM&  74.83&79.44& 81.71& 83.65& 83.99& 82.56& 83.20& 84.52& 85.23& 85.48& \textbf{86.72}&86.65\\
			\midrule
			& Base&  90.10&89.37& 90.70& 90.5& 89.80& 90.71& 90.32& \textbf{90.74}& 90.57& 90.13& 90.33&90.67\\
			Food101& New&  91.22&88.77& 91.29& 91.7& 91.37& \textbf{92.05}& 90.91& 91.85& 91.37& 91.30& 91.40&\textcolor{blue}{\textit{91.53}}\\
			& HM&  90.66&89.07& 90.99& 91.09& 90.58& \textbf{91.38}& 90.61& 91.29& 90.97& 90.71& 90.86&\textcolor{blue}{\textit{91.10}}\\
			\midrule
			& Base&  27.19&39.67& 33.41& 36.21& 42.13& 37.44& 47.08& 40.14& 41.97& 40.57& 45.97&\textbf{49.57}\\
			Aircraft& New&  36.29&31.23& 23.71& 33.55& 33.73& 35.61& 29.87& 36.51& 34.43& 36.33& 37.20&\textbf{37.63}\\
			& HM&  31.09&34.95& 27.74& 34.83& 37.46& 36.50& 36.55& 38.24& 37.83& 38.33& 41.12&\textbf{42.78}\\
			\midrule
			& Base&  69.36&80.85& 79.74& 80.29& 82.20& 80.82& 80.67& 82.26& 82.63& 82.27& \textbf{83.23}&83.00\\
			SUN397& New&  75.35&68.34& 76.86& 76.53& 73.63& 78.70& 76.11& 79.04& 78.20& 78.57& \textbf{79.20}&78.93\\
			& HM&  72.23&74.07& 78.27& 78.36& 77.68& 79.75& 78.32& 80.62& 80.35& 80.38& \textbf{81.17}&80.91\\
			\midrule
			& Base&  53.24&79.97& 77.01& 77.55& 81.97& 80.36& 83.95& 83.10& 82.77& 83.20& \textbf{85.63}&85.03\\
			DTD& New&  59.90&48.60& 56.00& 54.99& 43.80& 59.18& 59.06& 58.05& 58.07& \textbf{65.63}& 64.33&\textcolor{blue}{\textit{65.50}}\\
			& HM&  56.37&60.46& 64.85& 64.35& 57.09& 68.16& 69.34& 68.35& 68.25& 73.38& 73.47&\textbf{74.00}\\
			\midrule
			& Base&  56.48&90.10& 87.49& 85.64& 93.70& 94.07& \textbf{97.12}& 93.53& 91.63& 85.46& 95.93&\textcolor{blue}{\textit{96.67}}\\
			EuroSAT& New&  64.05&53.00& 60.04& 64.34& 62.67& 73.23& 72.91& 75.21& 74.73& 82.34& 71.10&\textbf{82.73}\\
			& HM&  60.03&66.74& 71.21& 73.48& 75.11& 82.30& 83.29& 83.38& 82.32& 83.87& 81.67&\textbf{88.66}\\
			\midrule
			& Base&  70.53&84.53& 82.33& 82.89& 86.60& 83.00& \textbf{88.56}& 85.33& 87.13& 86.23& 88.17&\textcolor{blue}{\textit{88.53}}\\
			UCF101& New&  77.50&67.37& 73.45& 76.67& 75.90& 78.66& 75.55& 77.72& 80.77& 80.03& 79.80&\textbf{79.97}\\
			& HM&  73.85&74.98& 77.67& 79.65& 80.90& 80.77& 81.54& 81.35& 83.83& 82.20& 83.78&\textbf{84.03}\\
			\bottomrule
	\end{tabular}}
	\caption{Comparison of FARL with previous methods on base-to-novel generalization. \textbf{Bold} indicates the best performance among all compared methods. Results shown in \textcolor{blue}{\textit{blue italics}} outperform previous SOTA method MMRL.}
	\label{tab:base2new}
	\vspace{-0.5cm}
\end{table*}

\section{Experiments}
\begin{figure}[!t]
	\centering
	\begin{minipage}[t]{0.32\textwidth}
		\centering
		\includegraphics[width=\linewidth]{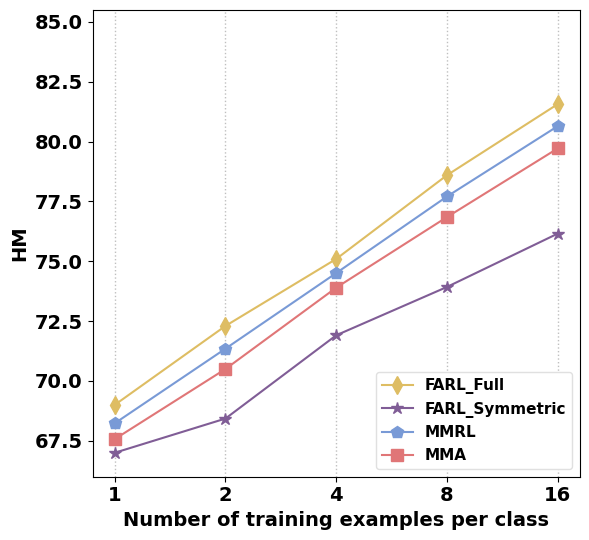}
		\caption{FARL and variants on few-shot learning across 11 datasets.}
		\label{fig:fewshot}
	\end{minipage}
	\hfill 
	\begin{minipage}[t]{0.29\textwidth}
		\centering
		\includegraphics[width=\linewidth]{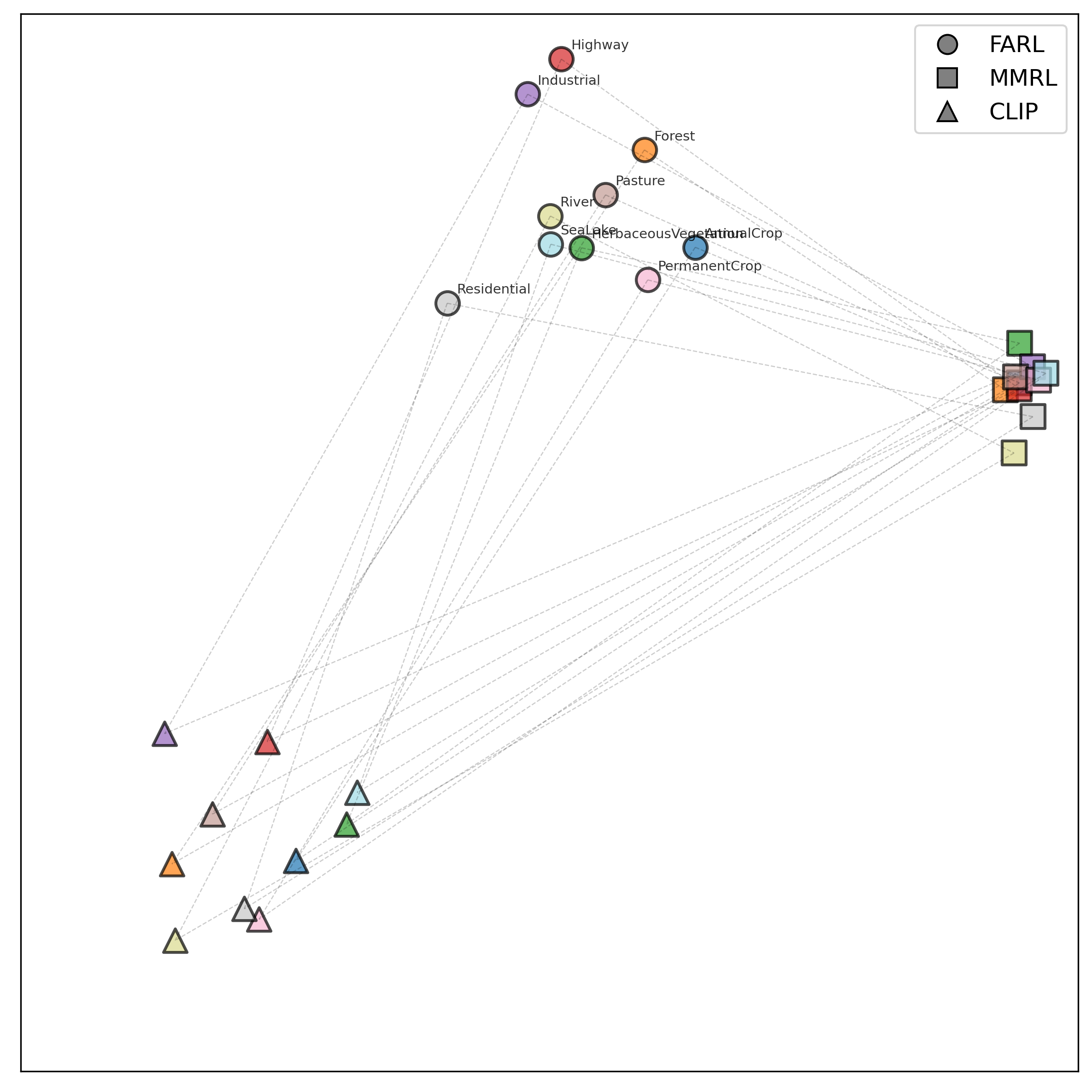}
		\caption{Compare text embeddings using PCA on EuroSAT.}
		\label{fig:eurosat_pca}
	\end{minipage}
	\hfill 
	\begin{minipage}[t]{0.29\textwidth}
		\centering
		\includegraphics[width=\linewidth]{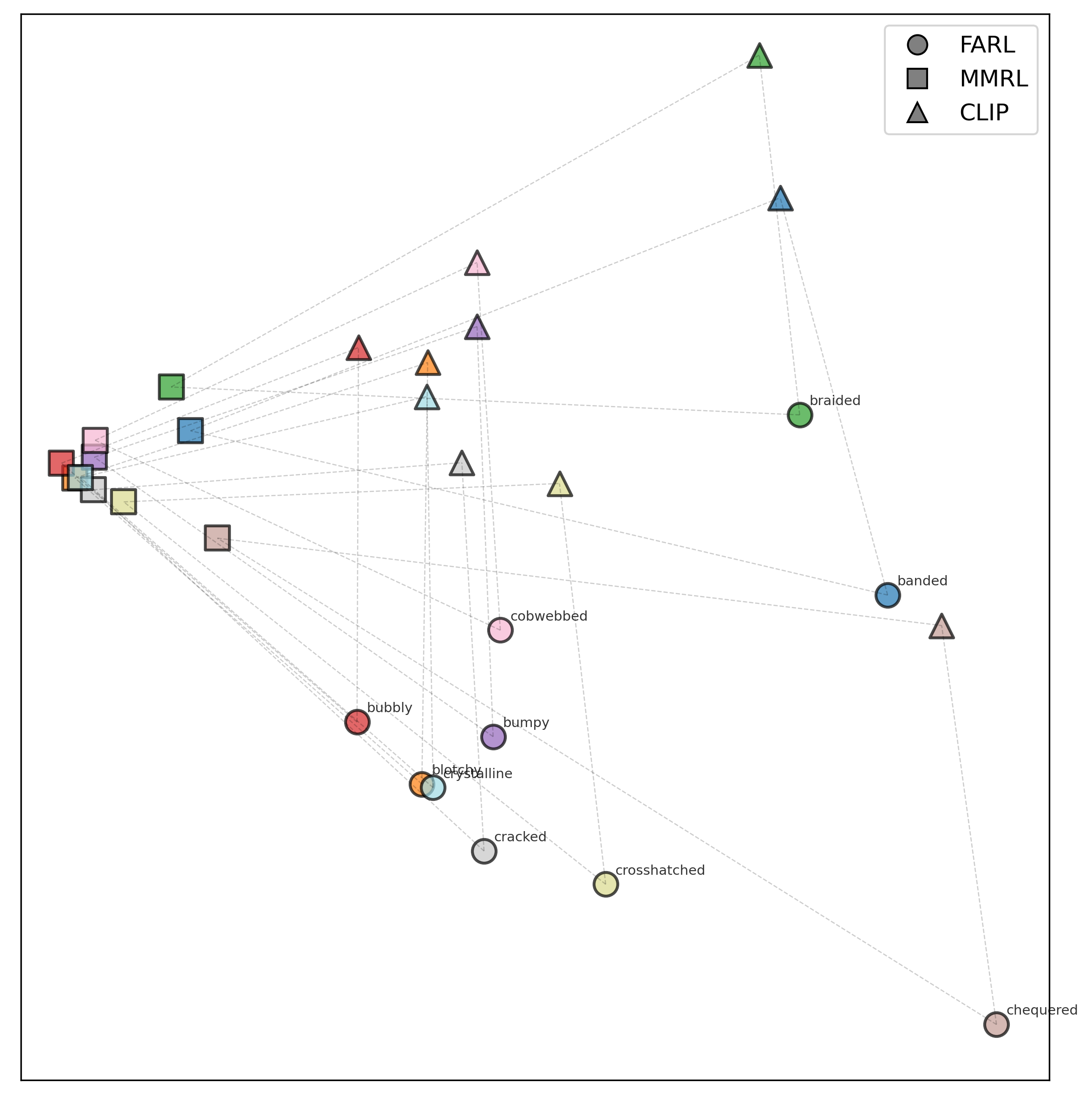}
		\caption{Compare text embeddings using PCA on DTD.}
		\label{fig:dtd_pca}
	\end{minipage}
	
	\vspace{-0.3cm}
\end{figure}

In this section, we conduct a series of experiments to evaluate our proposed Fourier-Attentive Representation Learning (FARL) framework against state-of-the-art methods across four standard benchmarks. Our goal is to validate FARL's effectiveness in few-shot adaptation, generalization to unseen concepts, and robustness to distribution shifts. In-depth analysis and ablation studies are presented in Section \ref{sec:ablation}.

\subsection{Experimental Setup}
Our evaluations are performed on a comprehensive suite of image classification datasets. For general-purpose tasks, we use 11 datasets: ImageNet\cite{imagenet}, Caltech101\cite{caltech101}, OxfordPets\cite{oxford_pets}, Flowers102\cite{flowers102}, Food101\cite{food101}, FGVCAircraft\cite{aircraft}, SUN397\cite{sun397}, DTD\cite{dtd}, EuroSAT\cite{eurosat}, StanfordCars\cite{stanford_cars}, and UCF101\cite{ucf102}. For the domain generalization task, we use ImageNet\cite{imagenet} as the source domain and evaluate on its four out-of-distribution variants: ImageNetV2\cite{imagenetv2}, ImageNet-Sketch\cite{imagenetsketch}, ImageNet-A\cite{imageneta}, and ImageNet-R\cite{imagenetr}.

We compare FARL against a range of strong and relevant baselines to position its performance, including: Zero-Shot CLIP\cite{clip}, prompt learning methods such as CoOp\cite{coop}, CoCoOp\cite{cocoop}, KgCoOp\cite{kgcoop}, PLOT\cite{plot}, ProVP\cite{provp}, MetaPrompt\cite{metaprompt}, MaPLe\cite{maple}, and TCP\cite{tcp}; and other deep-layer adaptation methods such as MMA\cite{mma} and MMRL\cite{mmrl}.

All models are built upon the ViT-B/16 CLIP backbone\cite{clip}. We follow the 16-shot learning protocol, where $16$ labeled examples per class are used for training. FARL are trained using the AdamW optimizer with an initial learning rate of $0.00005$ and a cosine decay schedule, we set the number of representation tokens $K=5$ and the injection start layer $J=6$. Detailed logic and pseudocode for the Fourier decomposition and model architecture are provided in the supplementary material. All results are the average of three runs with different random seeds. For a fair comparison, results for MMRL were obtained by re-running the official source code under our experimental setup.

\subsection{Base-to-Novel Generalization}
This is our primary evaluation to assess the trade-off between learning on seen classes and generalizing to unseen ones. For each dataset, the classes are split into two disjoint sets: base and novel. The model is trained only on the Base classes and evaluated on both. We report the accuracy on Base classes (measuring few-shot learning performance), accuracy on Novel classes (measuring zero-shot generalization), and the Harmonic Mean (HM) of the two as a balanced metric. Tabl. \ref{tab:base2new}.

\subsection{Cross-Dataset Evaluation}
To assess transferability, we train models on the 1000 classes of ImageNet and directly evaluate them on the other 10 datasets in a zero-shot manner. The results in Tabl. \ref{tab:crossdata} indicate that FARL maintains strong performance across diverse and unseen datasets, achieving the highest average accuracy. This highlights the transferability of the disentangled representations learned by our method.

\subsection{Domain Generalization}

We evaluate the robustness of the ImageNet-trained models on datasets with significant domain shifts (e.g., photos to sketches). As presented in Tabl. \ref{tab:domain_gen}, FARL demonstrates superior robustness to domain shifts compared to all baselines. We attribute this to the model's reliance on domain-invariant structural features extracted from the Fourier phase.

\begin{table}[!h]
	\centering
	\resizebox{\textwidth}{!}{
		\begin{tabular}{l|c|ccccccccccc} 
			\toprule\
			& \multicolumn{1}{c|}{Source} & \multicolumn{11}{c}{Target} \\
			\midrule
			& ImageNet & Average & Caltech101 & OxfordPets & StanfordCars & Flowers101 & Food101 & FGVCAircraft & SUN397 & DTD & EuroSAT & UCF101 \\
			\midrule
			CoOpOp \tiny{(CVPR2022)}   & 71.02 & 65.74 & 94.43 & 90.14 & 65.32 & 71.88 & 86.06 & 22.94 & 67.36 & 45.73 & 45.37 & 68.21 \\
			MaPLe \tiny{(CVPR2023)}    & 70.72 & 66.30 & 93.53 & 90.49 & 65.57 & 72.23 & 86.20 & 24.74 & 67.01 & 46.49 & 48.06 & 68.69 \\
			TCP \tiny{(CVPR2024)}      & 71.40 & 66.29 & 93.97 & 91.25 & 64.69 & 71.21 & \textbf{86.69} & 23.45 & 67.15 & 44.35 & \textbf{51.45} & 68.73 \\
			MMA \tiny{(CVPR2024)}      & 71.00 & 66.61 & 93.80 & 90.30 & \textbf{66.13} & 72.07 & 86.12 & 25.33 & \textbf{68.17} & \textbf{46.57} & 49.24 & 68.32 \\
			MMRL \tiny{(CVPR2025)}    & \textbf{73.70} & 66.13 & 94.43 & 91.30 & 64.67 & 71.87 & 85.10 & 25.37 & 67.00 & 44.03 & 48.60 & \textbf{68.93} \\
			\midrule
			FARL\tiny{(Ours)}          & 73.43 & \textbf{66.84} & \textbf{94.53} & \textbf{91.70} & \textcolor{blue}{\textit{66.07}} & \textbf{72.47} & \textcolor{blue}{\textit{86.07}} & \textbf{25.97} & \textcolor{blue}{\textit{67.53}} & \textcolor{blue}{\textit{44.60}} & \textcolor{blue}{\textit{50.67}} & 68.83 \\
			\bottomrule
		\end{tabular}%
	}
	\caption{Comparison of FARL with previous methods on cross-dataset evaluation.}
	\label{tab:crossdata}
	\vspace{-0.5cm}
\end{table}

\begin{table}[t!]
	\centering
	\begin{minipage}[t]{0.49\linewidth}
		\centering
		\caption{Comparison of FARL with previous methods on domain generalization.}
		\label{tab:domain_gen}
		\resizebox{\linewidth}{!}{%
			\begin{tabular}{l | c | c c c c}
				\toprule
				& \multicolumn{1}{c}{Source} & \multicolumn{4}{c}{Target} \\
				\cmidrule(lr){2-6}
				& ImageNet & -V2 & -S & -A & -R \\
				\midrule
				CLIP \tiny{(ICML2021)}     & 66.73 & 60.83 & 46.15 & 47.77 & 73.96 \\
				CoOpOp \tiny{(CVPR2022)}   & 71.02 & 64.07 & 48.75 & 50.63 & 76.18 \\
				MaPLe \tiny{(CVPR2023)}    & 70.72 & 64.07 & 49.15 & 50.90 & 76.98 \\
				MMA \tiny{(CVPR2024)}      & 71.00 & 64.33 & 49.13 & \textbf{51.12} & \textbf{77.32} \\
				MMRL \tiny{(CVPR2025)}     & \textbf{73.70} & \textbf{65.43} & 48.70 & 50.13 & 76.63 \\
				\midrule
				FARL\tiny{(Ours)}          & 73.43 & 64.83 & \textbf{49.23} & \textcolor{blue}{\textit{50.57}} & \textcolor{blue}{\textit{77.20}} \\
				\bottomrule
			\end{tabular}
		}
	\end{minipage}
	\hfill
	\begin{minipage}[t]{0.48\linewidth}
		\centering
		\caption{Ablation study on architectural variants.}
		\label{tab:variant}
		\resizebox{\linewidth}{!}{%
			\begin{tabular}{l|c|cc|c}
				\toprule
				Method & Architecture & Base & Novel & HM \\
				\midrule
				FARL & Phase + Amp & 86.11 & \textbf{77.49} & \textbf{81.57} \\
				\midrule
				$FARL_{Phase}$ & Phase Only & \textbf{86.14} & 76.82 & 81.21 \\
				$FARL_{Amp}$ & Amp Only & 83.86 & 73.05 & 78.08 \\
				\midrule
				$FARL_{Spatial}$ & RGB + RGB & 86.09 & 76.88 & 81.22 \\
				$FARL_{Phase+Spat}$ & Phase + RGB & 86.06 & 76.97 & 81.26 \\
				\midrule
				$FARL_{Symmetric}$ & Text \& Image & 83.14 & 70.26 & 76.16 \\
				\bottomrule
			\end{tabular}
		}
	\end{minipage}
\end{table}

\section{Analysis and Ablation Studies}
\label{sec:ablation}

\subsection{A Spectral View of Disentangled Adaptation}
\label{sec:spectral_view}

To understand why FARL generalizes better than holistic adaptation methods, we analyze the problem through a spectral lens. Let $x$ be an input image. Standard prompt learning methods optimize a function $f_\theta(x)$ (where $\theta$ are learnable prompts) to minimize empirical risk on a small support set $\mathcal{D}_{support}$.

\textbf{The Spectral Bias Hypothesis.} We argue that in the few-shot regime, $f_\theta$ is prone to a \textit{spectral bias}. The amplitude spectrum $\mathcal{A}(x)$ typically contains high-energy, low-frequency statistics (e.g., global lighting, background texture) that are "easier" for a neural network to fit than the phase spectrum $\mathcal{P}(x)$, which encodes complex spatial geometry (edges, shapes)~\cite{geirhos2022imagenettrainedcnnsbiasedtexture}.
When $|\mathcal{D}_{support}|$ is small, the model minimizes loss by latching onto $\mathcal{A}(x)$ correlations, which are often domain-specific. For example, if all "dogs" in the support set are on green grass, the model learns "green texture $\rightarrow$ dog" (an amplitude correlation) rather than "four-legged shape $\rightarrow$ dog" (a phase correlation). This leads to poor generalization on novel classes or domains where these amplitude statistics shift.

\textbf{FARL as Spectral Regularization.} FARL mitigates this by enforcing an explicit structural decomposition. Instead of a single holistic representation, we model the visual query $q$ as a weighted fusion of spectral components:
\begin{equation}
	q = w_p \cdot \psi(\mathcal{P}(x)) + w_a \cdot \phi(\mathcal{A}(x))
\end{equation}
where $\psi$ and $\phi$ are the phase and amplitude processing streams (via FFT and CNNs), and $w_p, w_a$ are dynamic attention weights learned by the cross-attention mechanism. This formulation is conceptual and used for analysis; in practice, the fusion is implemented via cross-attention and MLPs.

By explicitly separating these streams, FARL prevents the high-energy amplitude signal $\mathcal{A}(x)$ from drowning out the structural signal $\mathcal{P}(x)$ during the attention calculation.
\begin{itemize}
	\item \textbf{Phase Stream ($\psi$):} Acts as a \textit{Geometry Expert}. It forces the model to attend to spatial alignment and shape, which are robust invariant features essential for novel class recognition.
	\item \textbf{Amplitude Stream ($\phi$):} Acts as a \textit{Context Expert}. It captures energy distributions (texture, style). While less generalizable, it provides necessary context for base classes.
\end{itemize}
Crucially, the fusion mechanism allows the model to dynamically balance these cues: leaning on geometry for structure-rich classes (e.g., "Highway") while using context for texture-rich classes (e.g., "Fabric").

\subsection{Qualitative Analysis: Visualizing Disentanglement}

We validate this spectral hypothesis by visualizing the cross-attention maps, which indicate where the model focuses within each spectral stream.

\subsubsection{Object Recognition (OxfordPets):}
As shown in Fig.~\ref{fig:oxford_pets_fourier}, the attention maps reveal a clear separation of concerns:
\begin{itemize}
	\item \textbf{Phase Attention (Structure):} The model sharply focuses on high-frequency edges defining the cat's silhouette, ears, and facial contours. This confirms that the phase stream is successfully guiding the model to learn "what" the object is based on geometry.
	\item \textbf{Amplitude Attention (Style):} The attention is diffuse, covering the fur texture and background regions. This confirms the amplitude stream is capturing the "how" (style/lighting) without interfering with shape boundaries.
\end{itemize}

\subsubsection{Scene Recognition (EuroSAT):}
\begin{figure}[t]
	\centering
	\begin{subfigure}{\linewidth}
		\centering
		\includegraphics[width=0.19\linewidth]{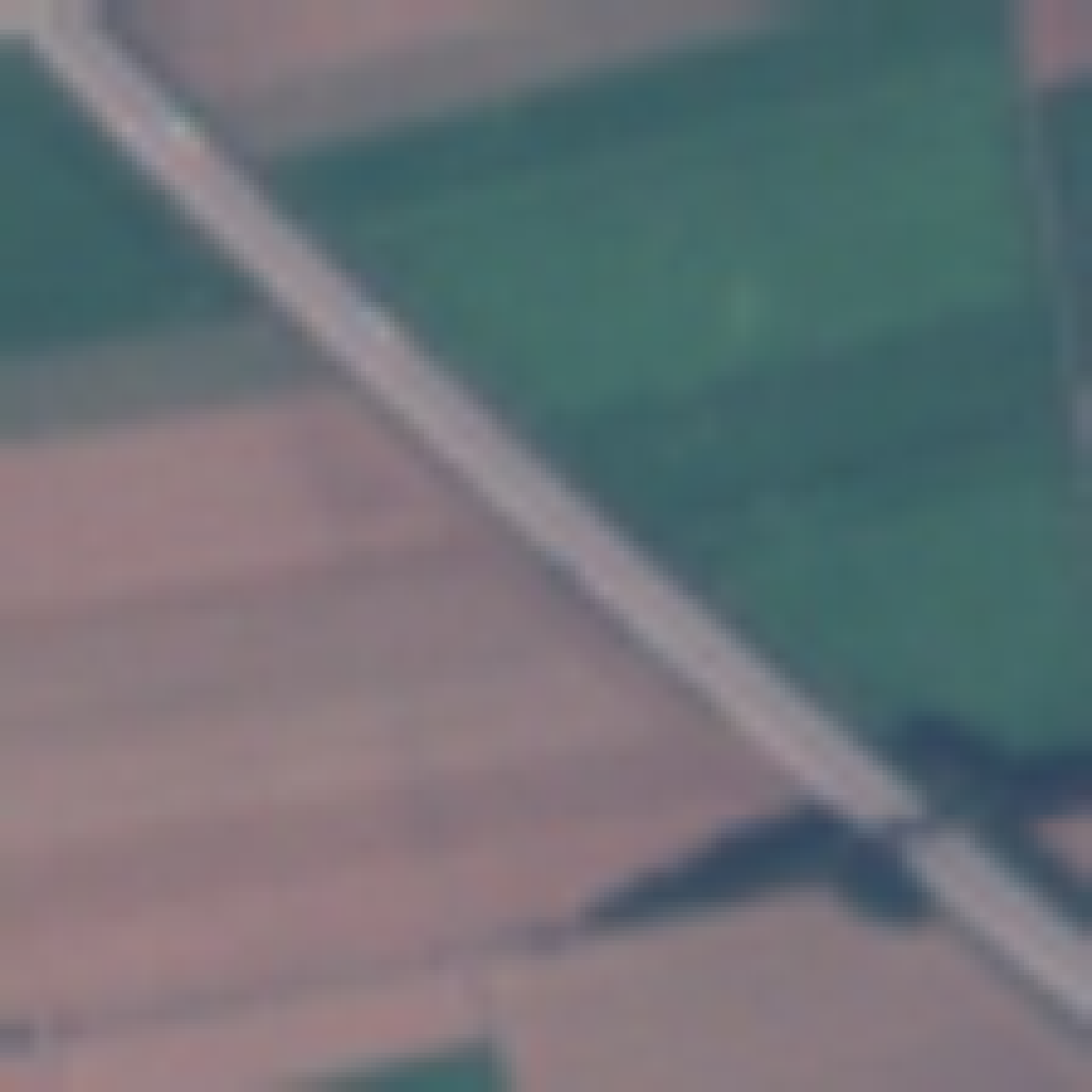}
		\includegraphics[width=0.19\linewidth]{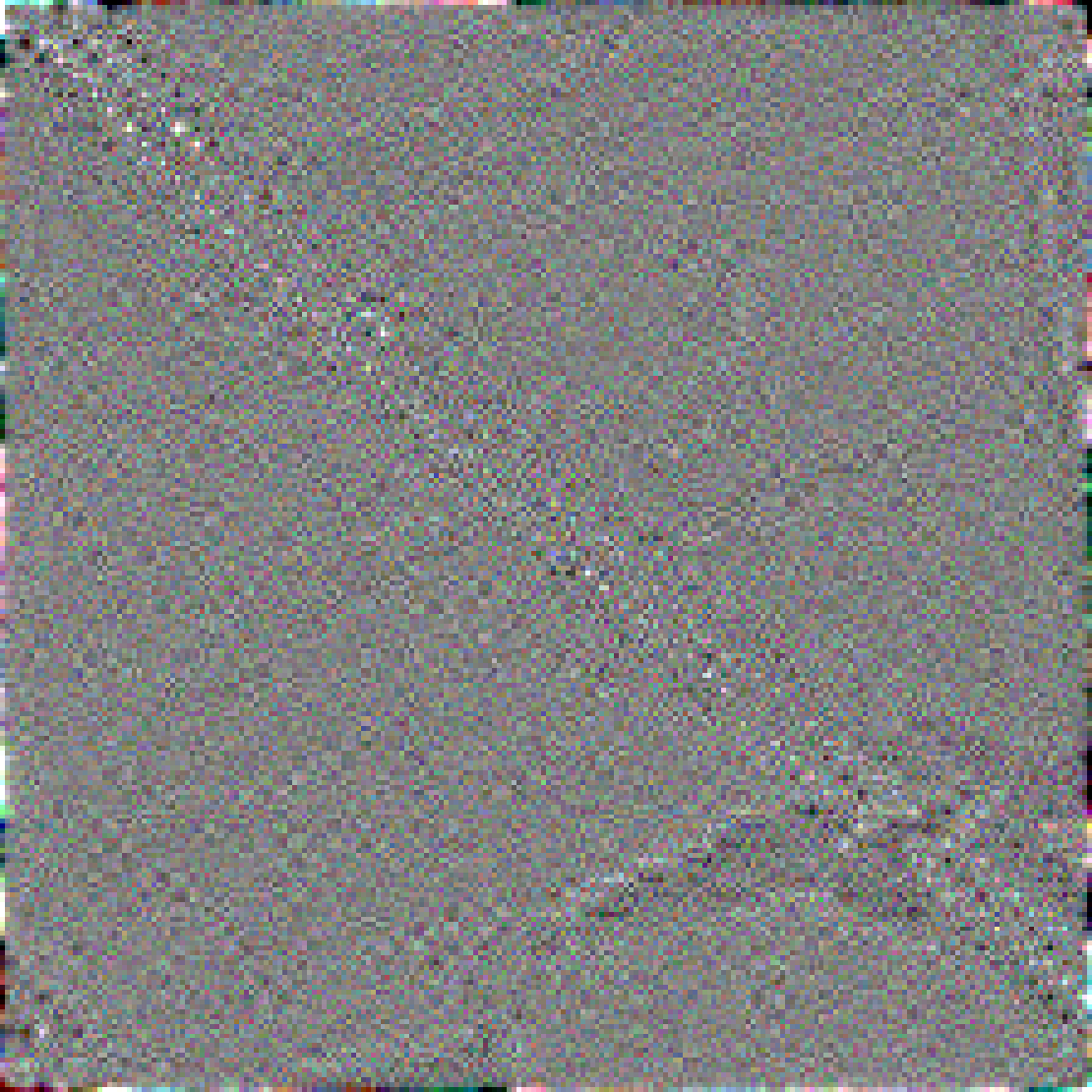}
		\includegraphics[width=0.19\linewidth]{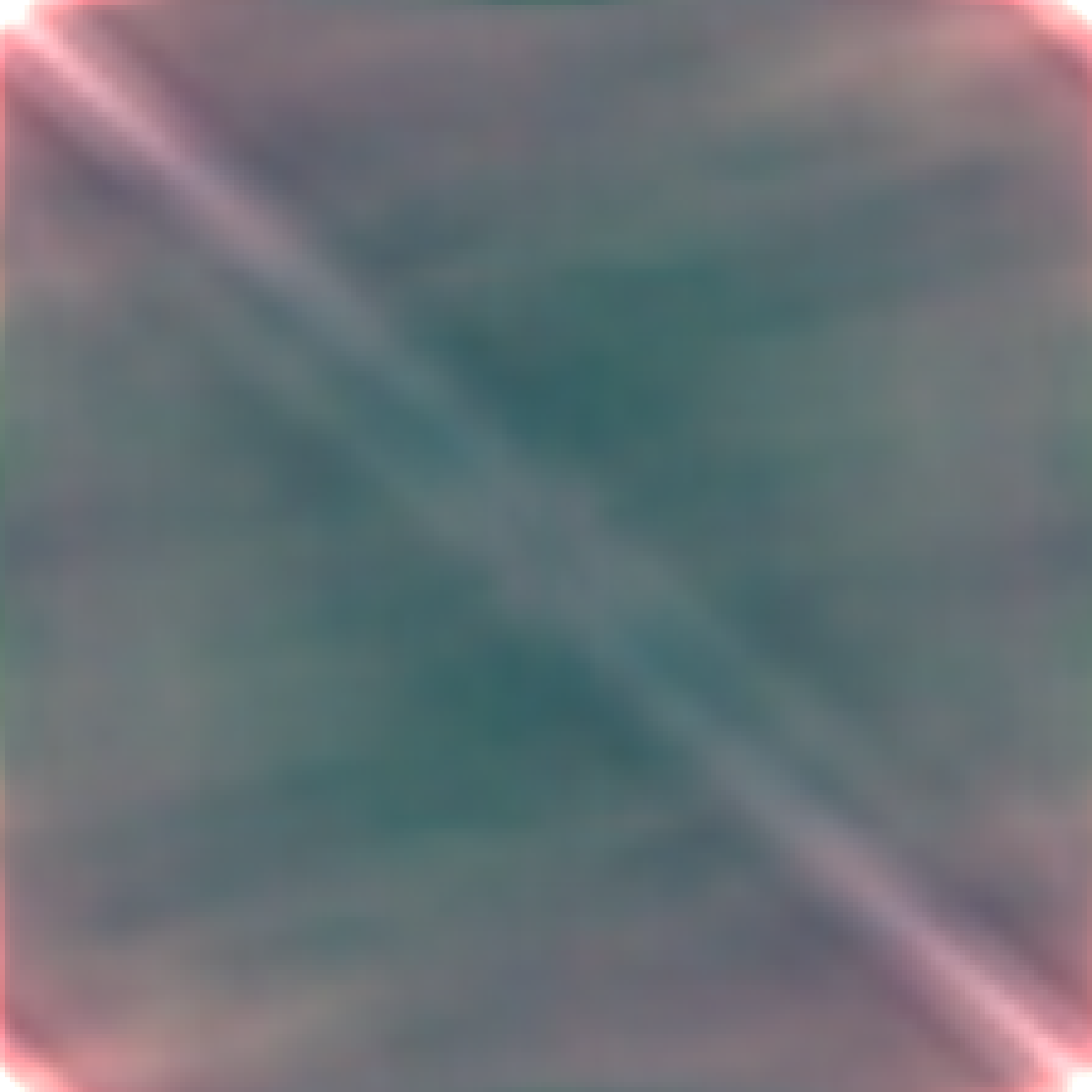}
		\includegraphics[width=0.19\linewidth]{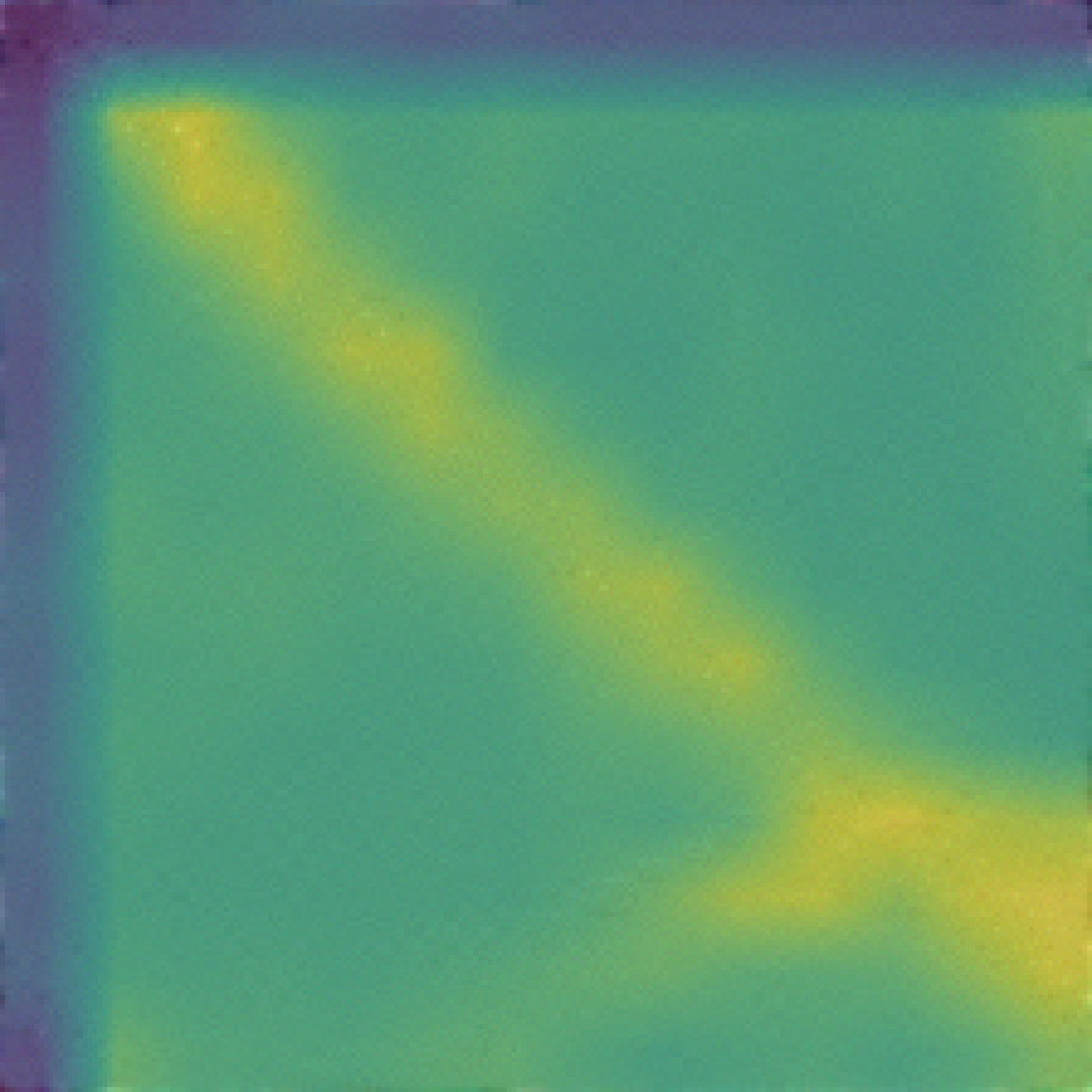}
		\includegraphics[width=0.19\linewidth]{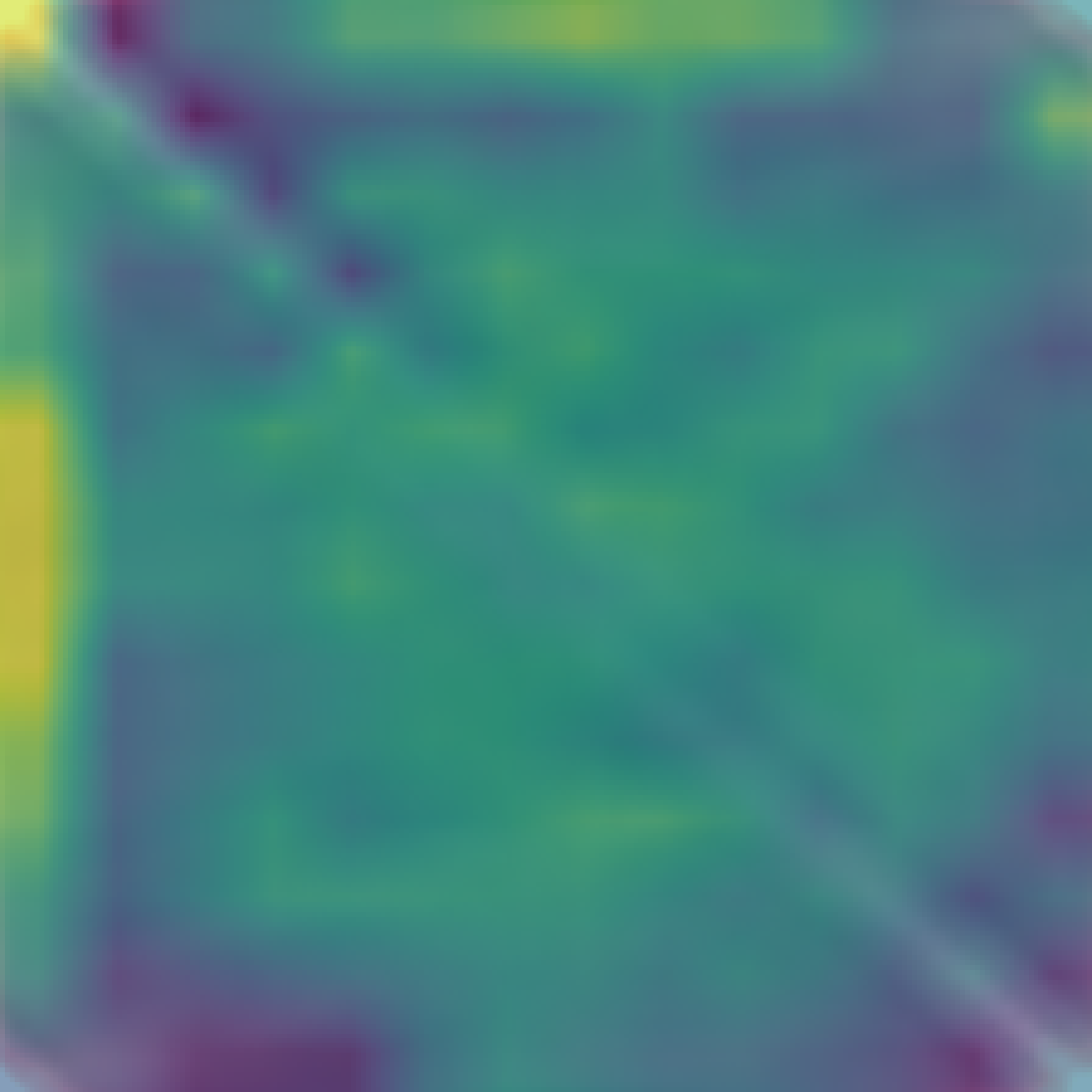}
	\end{subfigure}
	\caption{Visualizing spectral disentanglement on EuroSAT "Highway". \textbf{Phase attention} (4th col) isolates the road geometry, ignoring the surrounding vegetation texture captured by \textbf{Amplitude attention} (5th col).}
	\label{fig:eurosat_fourier}
	\vspace{-0.5cm}
\end{figure}

The impact is even more pronounced on the EuroSAT dataset (Fig.~\ref{fig:eurosat_fourier}), where classes like "Highway" are defined by geometric linearity rather than texture.
\begin{itemize}
	\item \textbf{Robustness to Texture Bias:} Standard models often confuse highways with rivers or herbaceous vegetation due to similar green textures (amplitude statistics).
	\item \textbf{Phase-Driven Distinction:} FARL's phase attention (Fig.~\ref{fig:eurosat_fourier}, 4th col) successfully ignores the green texture noise and locks onto the linear road structure. This ability to filter out amplitude noise explains FARL's superior performance (+10.03\%) on EuroSAT novel classes.
\end{itemize}

\subsubsection{PCA Embedding Analysis:}
To further validate spectral disentanglement, we visualize PCA projections of the learned embeddings on DTD and EuroSAT (Fig.~\ref{fig:dtd_pca}, \ref{fig:eurosat_pca}). 
Previous method MMRL\cite{mmrl} produce overlapping and elongated clusters, indicating entangled texture and structure cues. 
In contrast, FARL yields compact, well-separated clusters, especially on EuroSAT where classes differ mainly in global geometry (e.g., \emph{Highway} vs. \emph{River}). 
This suggests that explicit phase--amplitude decomposition reshapes the representation geometry by suppressing domain-specific amplitude bias and amplifying structure-aligned directions, directly supporting FARL’s superior few-shot generalization.

\subsection{Ablation Studies}
\label{variant_farl}
We investigate the contribution of each component to the spectral regularization effect. We compare the full FARL model against variants where streams are removed or replaced with raw spatial inputs Tabl. \ref{tab:variant}. Results are averaged over 11 datasets (16-shot).

\textbf{1. Phase Drives Generalization.} The $FARL_{Phase}$ variant performs nearly as well as the full model on novel classes (76.82\% vs 77.49\%). This supports our hypothesis that \textit{phase/structure is a major contributor of few-shot generalization}. Removing phase (i.e., $FARL_{Amp}$) leads to a massive drop in novel class accuracy (-4.44\%), proving that amplitude features alone are prone to overfitting.

\textbf{2. Amplitude Prevents Underfitting.} While phase is crucial for novelty, the full model outperforms $FARL_{Phase}$ slightly. This indicates that amplitude is not "noise" to be discarded; it provides complementary style information that helps disambiguate classes where structure is ambiguous (e.g., distinguishing "orange" from "lemon").

\textbf{3. Explicit Decomposition Matters.} A critical baseline is $FARL_{Spatial}$, which uses the same dual-stream architecture but feeds raw RGB images to both. It underperforms FARL (HM 81.22 vs 81.57). This proves that the gain does not come from simply adding parameters or branches. The \textit{explicit Fourier decomposition} is necessary to break the spectral bias and force the model to learn disentangled representations. Although the numerical gain appears modest, the improvement is consistent across datasets and crucially accompanied by clearer disentanglement, as evidenced by attention visualizations.

\section{Conclusion}
\label{sec:conclusion}

We revisited few-shot VLM adaptation through a spectral lens and identified \textbf{spectral bias}, the tendency to overfit domain-specific amplitude statistics, as a key obstacle to generalization. To address this issue, we proposed \textbf{FARL}, a framework that enforces representation-level spectral disentanglement beyond holistic feature tuning.

By explicitly separating phase-derived structure and amplitude-derived style and integrating them via an asymmetric injection strategy, FARL encourages models to prioritize robust geometric cues while retaining useful contextual information. Extensive experiments demonstrate improved robustness in base-to-novel generalization and cross-dataset transfer.

More broadly, our results suggest that integrating fundamental signal processing principles directly into the representation learning loop, rather than using them solely for data augmentation, offers a promising direction for few-shot vision-language adaptation.

%
%
\bibliographystyle{splncs04}
\bibliography{main}
\end{document}